%% file: main.tex
\documentclass[runningheads]{llncs}

\usepackage{eccv}

\usepackage{eccvabbrv}

\usepackage{graphicx}
\usepackage{booktabs}

\usepackage{listings}
\usepackage{xcolor}
\usepackage{multirow}
\usepackage{wrapfig}
\usepackage{float}
\colorlet{punct}{red!60!black}
\definecolor{background}{HTML}{EEEEEE}
\definecolor{delim}{RGB}{20,105,176}
\colorlet{numb}{magenta!60!black}

\lstdefinelanguage{json}{
    basicstyle=\small\ttfamily,
    numbers=left,
    numberstyle=\scriptsize,
    stepnumber=1,
    numbersep=8pt,
    showstringspaces=false,
    breaklines=true,
    frame=lines,
    backgroundcolor=\color{background},
    literate=
     *{0}{{{\color{numb}0}}}{1}
      {1}{{{\color{numb}1}}}{1}
      {2}{{{\color{numb}2}}}{1}
      {3}{{{\color{numb}3}}}{1}
      {4}{{{\color{numb}4}}}{1}
      {5}{{{\color{numb}5}}}{1}
      {6}{{{\color{numb}6}}}{1}
      {7}{{{\color{numb}7}}}{1}
      {8}{{{\color{numb}8}}}{1}
      {9}{{{\color{numb}9}}}{1}
      {:}{{{\color{punct}{:}}}}{1}
      {,}{{{\color{punct}{,}}}}{1}
      {\{}{{{\color{delim}{\{}}}}{1}
      {\}}{{{\color{delim}{\}}}}}{1}
      {[}{{{\color{delim}{[}}}}{1}
      {]}{{{\color{delim}{]}}}}{1},
}

\usepackage[accsupp]{axessibility}  

\usepackage{hyperref}

\usepackage{orcidlink}

\begin{document}

\title{Recommendation of data-free class-incremental learning algorithms by simulating future data} 

\titlerunning{Recommendation of DFCIL algorithms by simulating future data}

\author{Eva Feillet\inst{1,2}\orcidlink{0009-0007-4820-1599} \and
Adrian Popescu\inst{1}\orcidlink{0000-0002-8099-824X} \and
Céline Hudelot\inst{2}\orcidlink{0000-0003-3849-4133}}

\authorrunning{E.~Feillet et al.}

\institute{Université Paris-Saclay, CEA, LIST, F-91120, Palaiseau, France \\ 
Université Paris-Saclay, CentraleSupélec, MICS, France \\ 
\email{name.surname@\{cea.fr; centralesupelec.fr\}}}

\maketitle

\begin{abstract} 

Class-incremental learning deals with sequential data streams composed of batches of classes. 
Various algorithms have been proposed to address the challenging case where samples from past classes cannot be stored. 
However, selecting an appropriate algorithm for a user-defined setting is an open problem, as the relative performance of these algorithms depends on the incremental settings. 
To solve this problem, we introduce an algorithm recommendation method that simulates the future data stream. 
Given an initial set of classes, it leverages generative models to simulate future classes from the same visual domain. 
We evaluate recent algorithms on the simulated stream and recommend the one which performs best in the user-defined incremental setting. 
We illustrate the effectiveness of our method on three large datasets using six algorithms and six incremental settings.
Our method outperforms competitive baselines, and performance is close to that of an oracle choosing the best algorithm in each setting.
This work contributes to facilitate the practical deployment of incremental learning.

\end{abstract}

\input{sec/1_intro}
\input{sec/2_related}
\input{sec/3_pbstatement}
\input{sec/4_method}
\input{sec/5_datagen}
\input{sec/6_evalreco}
\input{sec/7_discussion}
\input{sec/8_conclusion}

%
%
\bibliographystyle{splncs04}
\bibliography{main}
\newpage
\appendix
\renewcommand{\thesection}{\Alph{section}}
\input{sec_supp/0_ref_datasets}
\input{sec_supp/1_text}
\input{sec_supp/2_image}
\input{sec_supp/3_efcil}
\input{sec_supp/4_results}

\end{document}

%% file: sec/1_intro.tex
\section{Introduction}
\label{sec_intro}
Continual learning (CL) aims at building models able to handle new data or tasks over time \cite{ring1994continual,parisi2019_continual}. 
Class-incremental learning (CIL), in which the data stream is composed of batches of classes~\cite{van2022three}, is an actively studied CL paradigm~\cite{belouadah2021_study,masana2022survey,parisi2019_continual}. At each step of a CIL process, the model is updated with a new batch of classes while attempting to maintain the performance on all previously learned classes. 
In many practical applications of CL, computational costs and memory budgets are important constraints \cite{hayes2022online,feillet2023advisil,harun2023efficient}. 
The data-free version of CIL (DFCIL) has recently gained attention because it requires lower storage~\cite{Hersche_2022_CVPR_constrained_FSCIL,srivastava2021resource,verwimp2023applications}, making it suitable for resource-constrained applications on embedded devices. It is also a relevant paradigm when training data cannot be stored for privacy reasons \cite{verma2023privacy,tabassum2021privacy}.
Recent comparative studies \cite{petit2024analysis,feillet2023advisil} have shown that among the DFCIL approaches proposed to date, no single is the best for all practical cases. The performance of DFCIL algorithms depends on the characteristics of the incremental process, e.g.,  the number of incremental steps, the number of classes per step, and the amount of training data available per class.
Given this performance variability, we aim to recommend an appropriate algorithm for a user-specified DFCIL scenario.
This recommendation problem has been addressed in~\cite{feillet2023advisil} using a large set of precomputed experiments run on reference datasets. 
While interesting, this method requires many precomputed experiments and does not account for the visual domain of the classification task. 

\begin{figure*}[!ht] 
    \centering\includegraphics[width=0.99\linewidth]{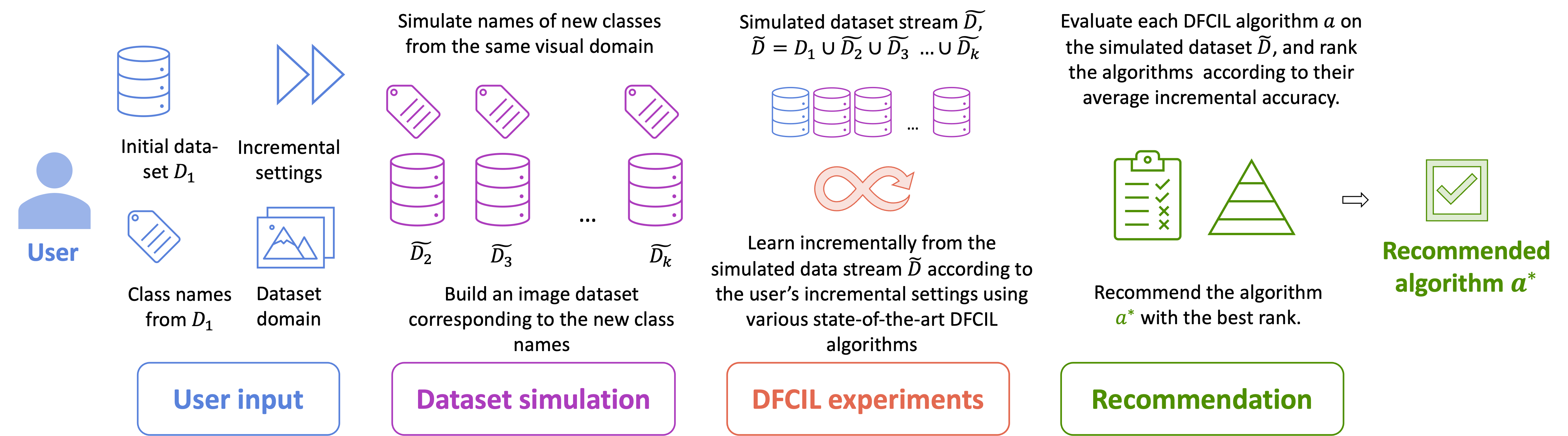} 
    \caption{Overview of the proposed method. A user needs an algorithm for a given DFCIL use case. 
    He has access to an initial labeled dataset $D_1$ and provides some expected characteristics of the incremental process, e.g. the number of classes per step. Based on these inputs, our method simulates a future data stream that extends $D_1$, first by proposing future class names, then by populating these classes with images. 
    Next, it evaluates different DFCIL algorithms on the simulated dataset and recommends the one with the best performance for deployment with real data.}
    \label{fig_principle}
\end{figure*}

In this article, we tackle the recommendation of DFCIL algorithms from a data-centric point of view.
Our method, illustrated in Figure~\ref{fig_principle}, takes as inputs the settings of the DFCIL process (number of incremental steps, number of classes per step) and a subset of classes available at the start of the process. 
Given a set of DFCIL algorithms, the recommended algorithm is obtained by: 
\begin{enumerate}
\item building a data stream that simulates future classes belonging to the same visual domain as the initial classes,
\item evaluating the candidate DFCIL algorithms on the simulated data stream, 
\item recommending the algorithm that performs best on the simulated stream.
\end{enumerate}

Our recommendation method is evaluated on three large datasets using six competitive DFCIL algorithms in six incremental scenarios. 
The results show that the performance of our method is close to that of an oracle that selects the best algorithm in any scenario. 
By simulating the future stream of data either using generative models or using the visual knowledge base ImageNet21k, our recommendation method compares favorably to the fixed choice of any of algorithms we tested, and AdvisIL~\cite{feillet2023advisil}. 
Additionally, we propose a strategy to lower the cost of exploring the performance of all candidate algorithms on the simulated data stream. 
We will share the code and data publicly.

%% file: sec/2_related.tex
\section{Related work}

\textbf{CIL algorithms.}
The ability to continually learn from new data is needed to develop more autonomous and sustainable systems \cite{ring1994continual,cossu2021sustainable}. 
One of the main challenges of CIL is to mitigate \textit{catastrophic forgetting} \cite{french1999catastrophic,kemker2018measuring}, namely the tendency of CIL models to abruptly forget previously acquired information when confronted with new information. 
To cope with this issue, numerous approaches have been proposed~\cite{parisi2019_continual,belouadah2021_study,masana2022survey}, based on 
(i) parameter-isolation \cite{aljundi2017_expertgate,rusu2016progressive,mallyaL18packnet},
(ii) iterative fine-tuning with distillation~\cite{li2016_lwf,hou2019_lucir,jodelet2023class,zhu2021pass,zhu2022self}, 
(iii) classifier-incremental learning with a fixed representation \cite{rebuffi2017_icarl,petit_2023_WACV,goswami2024fecam} 
and more recently, (iv) prompt-based methods relying on transformer architectures \cite{wang2022learning2prompt,jung2023instance_level_prompts}.

Catastrophic forgetting is particularly challenging in DFCIL, as optimizing the parameters of a deep neural network to recognize new classes without examples of past classes skews the classifier towards new classes~\cite{jodelet2021balanced}. 
Some DFCIL methods combine fine-tuning with knowledge distillation~\cite{hinton2015_distillation} to alleviate forgetting~\cite{sdc_2020,jodelet2021balanced,smith2021always,zhu2021class,zhu2021pass,zhu2022self}.
They handle new classes well since model parameters are updated to fit the novelty, but despite distillation, they tend to have lower performance for past classes~\cite{masana2022survey,zhu2022self}.
Another line of work proposes to use a fixed feature extractor trained during the initial step and focuses on incrementally learning only a classifier~\cite{hayes2020_deepslda,petit_2023_WACV,goswami2024fecam}.
The challenge is to leverage this fixed representation to separate all past and new classes well.
In our experiments, we include both types of methods to assess their strengths and limitations. 

\textbf{Generative models.}
We explore the potential of using generative models to simulate a future data stream. 
This stream is used to learn a classification model incrementally. 
Note that our goal is not to develop a competitor to existing generative models, but to use them in an innovative way for DFCIL.

In natural language processing, state-of-the-art large language models (LLMs) are based on generative transformer architectures \cite{brown2020language} that are trained in a self-supervised manner.
LLMs such as T5 \cite{raffel2020T5exploring}, Bloom \cite{scao2022bloom}, Llama-v2 \cite{touvron2023llama} or the family of GPT models \cite{brown2020language} show impressive results in generative tasks such as summarizing a text or writing a story. 
We use an LLM to simulate the future DFCIL process by generating class names related to the initial classes.

In computer vision, Variational Auto-Encoders \cite{Kingma2014VAEs} and Generative Adversarial Networks \cite{goodfellow2020generative} have been challenged by new algorithms and architectures, such as DALL-E~\cite{ramesh2021zero_DALL} and CLIP~\cite{radford2021clip}. 
These large multimodal models based on transformer architectures generate images from textual prompts.
Diffusion Probabilistic Models \cite{sohl2015deep_diffusion_probabilistic} (DPMs), based on a series of cascading denoising auto-encoders, now achieve state-of-the-art results.  
The popular Stable Diffusion (SD) method \cite{stablediff} improves the scalability of DPMs by performing denoising from a lower-dimensional space than the pixel space. SD models are designed with a general-purpose conditioning mechanism and can be prompted with textual descriptions to obtain high-quality images. 

Previous works~\cite{sariyildiz2023fake,tian2023stablerep} apply diffusion models to create synthetic datasets for transfer learning purposes. 
Image generation was used in continual learning but for generative replay \cite{shin2017generative_replay,van2021generative_replay,jodelet2023class}. 
Recently, the authors of \cite{jodelet2023class} proposed to generate images of past classes using SD at each step of the CIL process by prompting the generative model with their respective class names.  
This method achieves competitive performance but requires using a large diffusion model for the entire CIL process. 
This working hypothesis goes against the frugality requirements (memory, computation, latency) generally associated with CIL applications~\cite{prabhu2023computationally,cossu2021sustainable,belouadah2021_study}. 

In this article, we take inspiration from~\cite{sariyildiz2023fake} to simulate a dataset in the context of DFCIL. 
Unlike ~\cite{jodelet2023class}, \textit{}{we only use generative models in the initial non-incremental step of the CIL process}. 
This working hypothesis is compatible with CIL as data generation happens offline, for simulation only, and is not part of the incremental process.

\textbf{Simulating data streams for CL.} 
In~\cite{chrysakis2023simulating}, an algorithm is introduced to rearrange samples from a given dataset to form a data stream whose distribution changes continuously rather than in discrete steps as in batch-based CIL.
In~\cite{repetition2023}, a sampling-based generator creates arbitrarily long data streams with control over the repetition of past classes via probability distributions. 
These two approaches~\cite{chrysakis2023simulating,repetition2023} focus on evaluating streaming algorithms, whereas many DFCIL algorithms handle data that arrive in batches without repetition. 

The work that is closest to ours is AdvisIL~\cite{feillet2023advisil}. 
This method uses a set of precomputed DFCIL experiments done with auxiliary datasets to simulate incremental processes.
The main limitation of AdvisIL is the need for the user-defined DFCIL settings to be similar to the settings of a subset of the precomputed experiments. 
Furthermore, AdvisIL's recommendations do not consider the semantic characteristics of the incremental datasets, such as their visual domain or the granularity of the classification tasks. 

%% file: sec/3_pbstatement.tex
\section{DFCIL process}
In this section, we remind the DFCIL paradigm. 
We consider a dataset $\mathcal{D} = D_1 \cup D_2 \cup \dots \cup D_T$ and a sequential learning process composed of $T$ non-overlapping steps $s_1$,$s_2$,$\dots$,$s_T$. 
A step $s_i$ consists of learning from the labeled examples of the set of new classes $P_i$ from the subset $D_i$. 
Each sample from $D_i$ belongs to a unique class from $P_i$, and each class is present in a single data subset.
We denote by $n$ the average number of images per class in $D_1$. We consider that each of $P_2, P_3,\dots, P_T$ has the same number of classes $N$. 

\textbf{Model training.} At the first step $s_1$, the model $\mathcal{M}_1$ is trained on the data subset $D_1$ that involves the set of classes $P_1$. 
For each of the following steps, the same procedure is applied. 
For $i=2,3,\dots,T$, at the step $s_i$, the model $\mathcal{M}_i$ first recovers the parameters from the model $\mathcal{M}_{i-1}$ obtained in the previous step. 
It is then updated using the samples from $D_i$ to incorporate the new $P_i$ classes.
Depending on the DFCIL algorithm, only some of the model parameters are updated (e.g. only those of the classifier in \cite{hayes2020_deepslda,petit_2023_WACV,goswami2024fecam}, those of the classifier and some of the feature extractors in \cite{hayes2020_remind,petit2023plastil}, or all parameters in~\cite{li2016_lwf, hou2019_lucir, jodelet2021balanced}).
CIL algorithms should be designed for arbitrarily long incremental processes and any size of incremental batches.
In practice, for evaluation purposes, $T$ and $N$ are defined based on the datasets used in evaluation benchmarks~\cite{belouadah2021_study,masana2022survey,van2022three}. 

\textbf{Evaluation metric.} DFCIL algorithms are commonly evaluated based on their \textit{average incremental accuracy}, computed as $AA = \frac{1}{T} \sum_{i=1}^T q_i$, where $q_i$ is the accuracy of the model $\mathcal{M}_i$ on test samples from $\bigcup_{j=1}^i D_j$, after learning at step $s_i$. 

%% file: sec/4_method.tex
\section{Method} \vspace{-0.6em}
We introduce a method that recommends a DFCIL algorithm according to the characteristics of the incremental process and an initial dataset $D_1$. 
We explain our working hypotheses in Subsection \ref{subsec_hyp}.
The first step of our method is to build a simulated dataset that will be used as a proxy for the future data stream. 
In Subsection~\ref{subsec_gen}, we present two approaches for building such a simulated dataset. 
The first approach relies on a language model to generate new class names along with a description of each class, to extend $D_1$. 
We then populate each new class by prompting a text-to-image model with its name and description. 
The second approach queries a visual knowledge base to form a data stream that is semantically similar to the original class names.
Finally, we present in Subsection \ref{subsec_reco} how to recommend a DFCIL algorithm by evaluating candidate algorithms on the simulated dataset.

\vspace{-1.2em}
\subsection{Working hypotheses}
\label{subsec_hyp}
\vspace{-0.2em}

We describe the main working hypotheses made in this work. 
The first one relates to the characteristics of the incremental process.
While our method can be applied in any setting, it also inherits the evaluation-related constraints of the algorithm it compares~\cite{belouadah2021_study,masana2022survey,van2022three}. Thus, in addition to $D_1$, we assume that the user provides an estimate of the number of classes per step $N$ and an estimate of the number of incremental steps $T$.

Second, we make the usual supervised learning assumption that the class labels from $P_1$ are known. 
Confirming the results of~\cite{sariyildiz2023fake}, we found in preliminary experiments that enriching prompts with visual descriptions of the classes improves the conditioning of the generation of images with Stable Diffusion. 
These class descriptions are either generated automatically by an LLM or readily available when using a preexisting visual knowledge base.

Third, following existing works~\cite{belouadah2021_study,castro2018_e2eil,hayes2020_deepslda,hayes2020_remind}, we distinguish between the first \textit{non-incremental} step $s_1$ and the following steps and consider that compute and memory constraints apply after the initial step.
Under this assumption, we use generative models to simulate the future stream and to recommend a suitable algorithm before deployment in a user-specified scenario.
Importantly, in the illustration of our recommendation method in Section \ref{subsec_reco_results}, we select DFCIL algorithms with comparable inference memory requirements to ensure a fair comparison.

Finally, following common CIL benchmarks using ImageNet subsets~\cite{deng2009_imagenet,hendrycks2021many,goswami2024fecam} or more fine-grained thematic datasets (see examples in~\cite{masana2022survey}), we assume that the visual domain of the incremental dataset does not abruptly change over time, e.g. if the initial classes represent landmarks, we simulate future classes which also represent landmarks. 
Our data simulation method is not bound to this hypothesis, i.e. the generative models can be prompted to introduce a domain shift if the user wants to, but we make this choice to circumscribe our evaluation.

\subsection{Building simulated datasets}
\label{subsec_gen}
We denote a simulated dataset by $\mathcal{\Tilde{D}} = \Tilde{D}_1 \cup \Tilde{D}_2 \cup \dots \cup \Tilde{D}_T$, 
 where $\Tilde{D}_1 = D_1$ (so $\Tilde{P}_1 = P_1$) and for $i \geq 2$, $\text{Card}(\Tilde{P}_i)=N$ and $\text{Card}(\Tilde{D}_i)=N\cdot n$. 
The simulated dataset $\mathcal{\Tilde{D}}$ satisfies the following constraint: for $i,j \in \{1,2,\dots, T\}$ with $i \neq j$, $\Tilde{P}_i \cap \Tilde{P}_j = \emptyset$ (no repetition of a class across the steps).
In the following, we describe two approaches to obtaining $\mathcal{\Tilde{D}}$, which will be a potential future stream with classes from the same visual domain as $\mathcal{D}_1$. 
To control the semantic content of the simulated stream, each approach begins by constructing a set of $(T-1)\cdot N$ new class names $\Tilde{P}_2 \cup \Tilde{P}_3 \cup \dots \cup \Tilde{P}_T$, ensuring that these new classes do not appear in $P_1$. 
Then, each class is populated with either generated or real images. 

We performed preliminary experiments that simulate future classes using geometric data augmentation, such as rotations or mixing up class pairs.
However, the simulated data streams were too far from the real data distribution to obtain relevant recommendations (more details in the supplementary material). 

\vspace{-1.2em}
\subsubsection{Generative simulation}
Our first approach, named \textbf{SimuGen}, uses an LLM and a text-to-image model. 
The LLM automatically provides visual descriptions for the class names from $P_1$ when not readily available from a resource such as WordNet~\cite{miller1995wordnet}.
Then, building on the representation of the visual task provided by the initial class names and their descriptions, we aim to obtain $(T-1) \cdot N$ classes within the same visual domain as $D_1$. 
We build a list $L$ of pairs of the form $(c,d)$, where $c$ is the name of a class in $P_1$ and $d$ is its associated description. 
An LLM can produce many different class names, and the prompts can be advantageously tweaked to obtain class names more or less similar to those in $P_1$. 
In practice, we choose LLamav2-13b-chat~\cite{touvron2023llama} as it balances performance and inference time.
From a sublist of length 3 of $L$ and the visual domain of the initial dataset, we prompt the language model with the following pattern:

\vspace{1mm}
\noindent{\begin{minipage}{0.999\textwidth}
\small ``\texttt{Here is a list of [visual domain]: [sublist]. Could you provide ten more items on the same topic, with a short visual description of each item?}''. 
\end{minipage}}
\vspace{1mm} 
\\
The textual output of the LLM allows us to form a new list of pairs $(c', d')$, with $c'$ a new class name and $d'$ its associated description.
In the prompt, the description $d$ facilitates disambiguation. 
We also observed that asking for a visual description of the items produced more relevant suggestions for class names.
Since the input and output lengths of the LLM are limited, we prompt it multiple times with different sublists instead of once with the entire list, and we ask for ten new items at each time. 
Diversifying the sublists also diversifies the results. 
We iterate the process until we obtain $N \cdot (T-1)$ new unique class names to extend the initial subset of class names $P_1$. 
To facilitate postprocessing, we use a system prompt asking for a JSON output.
We provide more details about the use of LLamav2-13b-chat in the supplementary.

The second step of SimuGen consists of generating $n$ examples for each new class name. 
We note that generative text-to-image models based on Stable Diffusion~\cite{stablediff} have recently been used to generate visual datasets whose transfer learning performance is on par with or even superior to that of real datasets~\cite{sariyildiz2023fake,tian2023stablerep}. 
We use the Stable-Diffusion-2-1-base version that provides high-quality images.
For a given class, we obtain its associated images by prompting the model with its name $c'$ and its short visual description $d'$ obtained in the first step of SimuGen. As reported in~\cite{sariyildiz2023fake}, associating class names with some context produces better image diversity and avoids the pitfalls of rare or ambiguous words.
We use the following pattern: \small
``\texttt{a [style] photo of a [class name], [description]}", \normalsize
where style is picked from a small dataset-related list.
A prompt example used to generate an image is 
\small``\texttt{a panorama photo of Salar de Uyuni, the world's largest salt flat, Bolivia}". \normalsize
The process is repeated with different random seeds until $n$ images are obtained for each new class. 
\vspace{-0.4em}

\subsubsection{Simulation using a knowledge base}
Our second approach, named \textbf{Proxy21k}, selects new classes from an existing large-scale and general-purpose visual dataset built on top of a knowledge base.  
In our experiments, we use the ImageNet dataset~\cite{deng2009_imagenet}, which populates a subset of the WordNet lexical database~\cite{miller1995wordnet} with images.
ImageNet has a hierarchical structure and includes over 21,000 classes.
We prune ImageNet to keep only subtrees of the visual database related to the domain of the initial dataset. 
Then, we randomly pick $N \cdot (T-1)$ new classes among the ImageNet leaf classes from the preselected subtrees having at least $n$ images.
One underlying assumption of Proxy21k is that ImageNet-21k allows the sampling of a sufficiently big simulated dataset that covers the same visual domain as $\mathcal{D}$. 
This assumption is strong when ImageNet-21k does not sufficiently cover the target visual task. 

\subsection{Recommending a DFCIL algorithm}
\label{subsec_reco}

Let $\mathcal{A}$ be a set of candidate DFCIL algorithms. We consider that a simulated dataset $\mathcal{\Tilde{D}}$ was created using either the SimuGen or Proxy21k approach. 
We consider three recommendation strategies that use $\mathcal{\Tilde{D}}$. 

\subsubsection{Greedy recommendation}
This strategy consists in evaluating the performance of each algorithm in $\mathcal{A}$ on the simulated dataset $\mathcal{\Tilde{D}}$ and then recommending the DFCIL algorithm with the best average incremental accuracy on $\mathcal{\Tilde{D}}$ (see results in Subsection \ref{subsec_reco_results}). 

\subsubsection{Efficient simulation}
To limit the computational cost of the DFCIL experiments, we consider two recommendation strategies based on a partial simulation of the future stream:\\
(i) ``$t$-greedy exploration'' runs all algorithms in $\mathcal{A}$ during $t$ steps, then recommends the algorithm with the best average performance. \\
(ii) ``Explore then prune'' runs all algorithms in $\mathcal{A}$ during $t$ steps, then keeps exploring until step $t'$ while eliminating the worst-performing candidate algorithm at each step. The algorithm with the best average performance among the remaining candidates is recommended.

We remind that like AdvisIL~\cite{feillet2023advisil}, our method recommends a DFCIL algorithm adapted to user-provided incremental learning settings.
Unlike AdvisIL, we adopt a data-centric point of view to personalize the recommendation: (i) we try to take into account the semantic content of the user's dataset, and (ii) our recommendation method is not based on a set of precomputed experiments, so it can provide relevant recommendations whatever the incremental settings.  

%% file: sec/5_datagen.tex
\section{Comparison of simulated datasets}
In this section, we apply SimuGen and Proxy21k to three reference datasets covering different visual domains and analyze the characteristics of the obtained simulated datasets. 

\subsection{Reference datasets}
\label{subsec_datasets}
We experiment with the following reference datasets: ILSVRC~\cite{olga2015_ilsvrc}, iNaturalist 2018~\cite{van2018inaturalist}, and Google Landmarks v2~\cite{noh2017_landmarks}.
We sample three balanced 1000-class subsets denoted IN1k, iNat1k, and Land1k, with 350, 310, and 330 images per class, respectively.
These datasets cover diversified visual tasks and allow us to assess the advantages and limitations of the proposed simulation approaches.

\subsection{Analysis of simulated class names}
\label{subsec_comp_class_names}
\begin{figure*}[b] 
 \centering 
 \subfloat[IN1k]{%
      \includegraphics[width=0.3\textwidth]{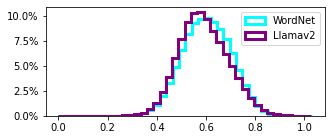} \label{fig_ilsvrc_hist}
      }
 \quad
 \subfloat[iNat1k]{%
      \includegraphics[width=0.3\textwidth]{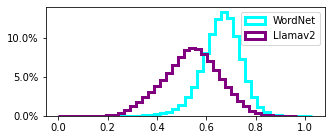}   \label{fig_inat_hist}}
 \quad
 \subfloat[Land1k]{%
      \includegraphics[width=0.3\textwidth]{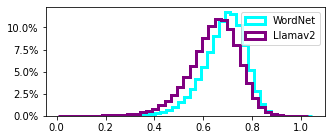}} 
\caption{Average cosine distances between CLIP embeddings of new class names obtained using either \color{cyan}Proxy21k (WordNet) \color{black} or \color{violet}SimuGen (Llamav2) \color{black} and the real class names of each dataset. Lower distances indicate a better fit between simulated and real class names.}%
 \label{fig_histograms}
\end{figure*}

In the following, we compare the actual class names from IN1k, iNat1k, and Land1k and their simulated counterparts obtained with SimuGen and Proxy21k, respectively. 
The goal is to understand how well each simulation approach introduced in Subsection \ref{subsec_gen} approximates the data stream.
To apply Proxy21K, we select classes from ImageNet by randomly sampling 1,000 classes as follows:
IN1k - all leaf classes to simulate generic fine-grained classification; 
iNat1k - leaf classes subsumed by the concepts ``flora'', ``fauna'' and ``fungi'' to cover natural concepts; 
Land1k -  leaf classes subsumed by Google Landmarks v2 categories, such as ``mountain'', ``museum'', or ``market'' (see the complete list in the supplementary).

We compute the CLIP embeddings \cite{radford2021clip} of the class names. 
The CLIP pretrained model projects textual and visual content in the same embedding space. 
Using CLIP embeddings of class names enables an indirect comparison of their associated images, all the more so as we used the CLIP text encoder to condition the image generation with Stable Diffusion.
In Figure~\ref{fig_histograms}, for each dataset, we compute the average cosine distance between the embedding of each class name in the dataset and the embeddings of the new class names obtained with SimuGen (\color{violet}purple curves\color{black}) and with Proxy21k (\color{cyan}blue curves\color{black}). 
These distributions show that SimuGen (Llamav2-13b-chat) offers a better domain approximation of the reference datasets than Proxy21k (WordNet). This applies to IN1k, even if, as ILSVRC class names are all covered by WordNet, we expected the class names obtained with Proxy21k to be closer to the real ILSVRC names. 
We note that the distributions of distances obtained with SimuGen and Proxy21k are more similar for ILSVRC and Land1K compared to iNat1k. This is partly because the LLM generates class names with scientific (Latin) names of the natural species, like the original class names from iNat1k, whereas scientific names are not always provided in WordNet.
Further illustrations are provided in the supplementary, including UMAP projections of class names.

\subsection{Analysis of generated images}
\label{subsec_comp_images}
\begin{figure}
    \centering 
    \includegraphics[width=0.75\linewidth]{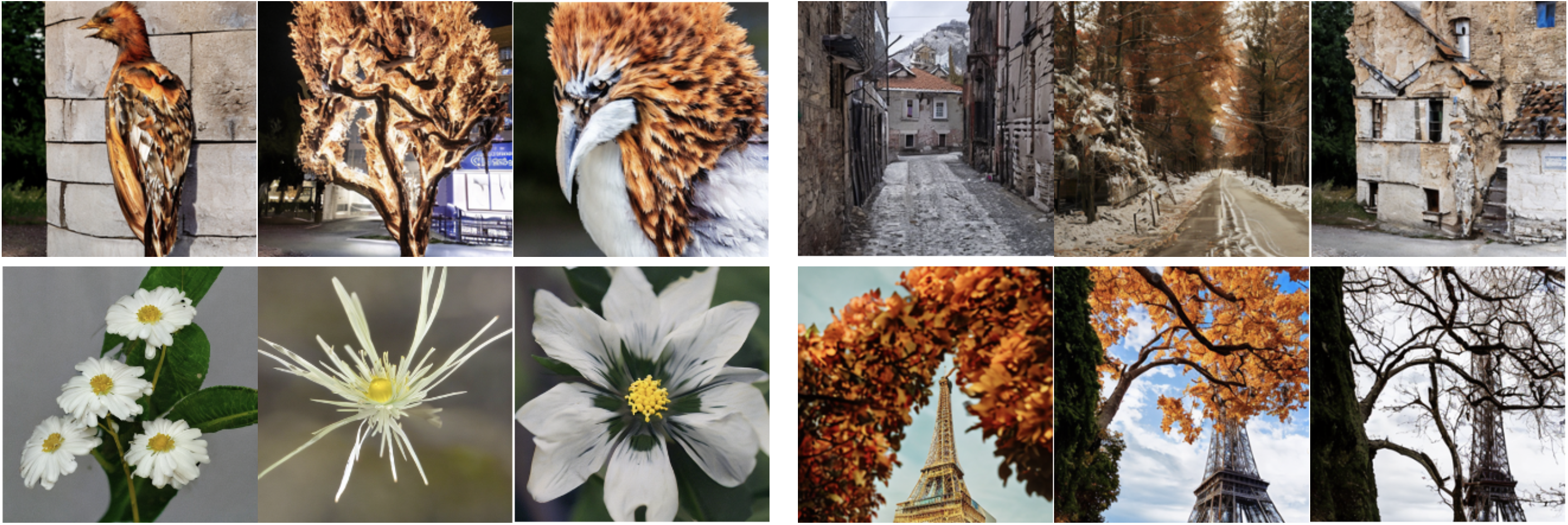} 
    \caption{Examples of images generated by Stable-Diffusion-v2-1-base with SimuGen (same prompt for each group, but different random seeds).}
    \label{fig_sd_examples} 
\end{figure}

For each dataset, following the Proxy21k approach, we sampled $n$ images from the ImageNet21k class corresponding to each selected WordNet synset, and following the SimuGen approach, we prompted Stable Diffusion-2-1-base~\cite{stablediff} with different seeds to obtain $n$ generated images per class.
We used SD to obtain large datasets flexibly, with control over the semantic content of the data and without the need for manual curation (unlike web crawling or using the unlabeled training corpus of LAION \cite{schuhmann2021laion} on which SD is trained).
In Figure \ref{fig_sd_examples}, we highlight that the images generated with SimuGen have a satisfying visual quality and are generally relevant to the class names.
This supports our use of a brief description of the class name in LLM prompts. 
Nonetheless, the diversity of the images of a given class could still be improved. 
On the one hand, some classes exhibit a low visual diversity, with well-known places such as the Eiffel Tower being depicted in a stereotyped way. 
On the other hand, rare class names (``Lycopus uniflorus''), imaginary concepts (``phoenix''), or polysemous words (``Bistritsa'' which can refer to a river, a village, etc.) may be depicted inaccurately or with a lot of visual diversity from one random seed to another. 
Further illustrations are provided in the supplementary.

%% file: sec/6_evalreco.tex
\section{Results}
\subsection{Evaluation framework}
\subsubsection{DFCIL algorithms}

We consider a set $\mathcal{A}$ composed of six candidate DFCIL algorithms.
Four of them rely on a fixed feature extractor and learn new classifiers incrementally.
NCM~\cite{rebuffi2017_icarl} uses a nearest-class mean classifier, DSLDA~\cite{hayes2020_deepslda} a streaming LDA, FeTrIL~\cite{petit_2023_WACV} linear SVCs~\cite{fabian2012_scikitlearn}, and FeCAM~\cite{goswami2024fecam} a Bayesian classifier based on the Mahalanobis distance. 
PlaStIL~\cite{petit2023plastil} freezes a part of the backbone and combines fine-tuning of the last layers with linear SVCs.
BSIL~\cite{jodelet2021balanced} is a fine-tuning-based algorithm that relies on a weighted softmax loss to rebalance predictions between old and new classes.
Note that we performed preliminary experiments (on Land1k) with other fine-tuning-based methods, namely SDC\cite{sdc_2020}, IL2A\cite{zhu2021more}, and PASS~\cite{zhu2021pass} but they were not recommended in any of the tested settings, confirming previous results reported in~\cite{goswami2024fecam,petit_2023_WACV}.
After also considering the high computational cost of training SDC, PASS, and IL2A, we did not retain them for the main experiments since their inclusion would not change the findings of this work. 
Additionally, we could not scale ABD~\cite{smith2021always} to 1,000 classes due to an inherent limitation of its model inversion component. 

To ensure comparability, all algorithms are implemented with the ResNet18 architecture~\cite{he2016_resnet} commonly used in CIL~\cite{castro2018_e2eil,hou2019_lucir,rebuffi2017_icarl}. Given a dataset and an initial number of classes, all algorithms share the same initial model (trained with the BSIL code). 
Note that for a fair comparison regarding the amount of data stored, we use the version of FeCAM that stores a single covariance matrix for all classes. Please refer to the supplementary material for further implementation details. 

\subsubsection{DFCIL scenarios}
The six candidate algorithms are evaluated in six incremental scenarios that push them to their limits. 
Three scenarios follow the protocol of~\cite{rebuffi2017_icarl} in the case of a 1000-class dataset: 50 steps of 20 classes each, 10 steps of 100 classes each, or 5 steps of 200 classes each. 
The other scenarios follow the protocol of~\cite{hou2019_lucir}, where the initial dataset contains half of all classes.
The remaining classes are split into 5, 10, or 100 steps (of 100, 50 or 5 classes respectively). 

\subsubsection{Performance}
We run each algorithm in each incremental scenario on each real dataset (IN1k, iNat1k, and Land1k) and their corresponding simulated datasets obtained with SimuGen and Proxy21k.
We denote by $\rho_{ref}$ the oracle method that selects the algorithm which performs best on the real dataset. 
We aim for the recommendation to behave like $\rho_{ref}$.
We denote by $\rho_m$ the accuracy obtained with the recommendation method $m$ ($gen, proxy, adv$ for SimuGen, Proxy21k, and AdvisIL, respectively) or the fixed use of a candidate algorithm. 
$\Delta_{m}^{s}$ denotes the gap between the average incremental accuracy of the algorithm provided by the oracle $\rho_{ref}$ and that of the algorithm recommended by the method~$\rho_m$, when the strategy $s$ is applied (``$T$'' when simulating all $T$ steps, ``$3$'' when simulating only the 3 first steps, ``effi'' when simulating 3 steps then discarding the worst performing algorithm until only one remains, or $T$ is reached). As the fixed baselines do not depend on the number of steps, we omit their superscript.

\subsection{Main results}
\label{subsec_reco_results}
\begin{figure}[t] 
 \centering
 \subfloat[Land1k]{
      \includegraphics[height=0.18\textheight]{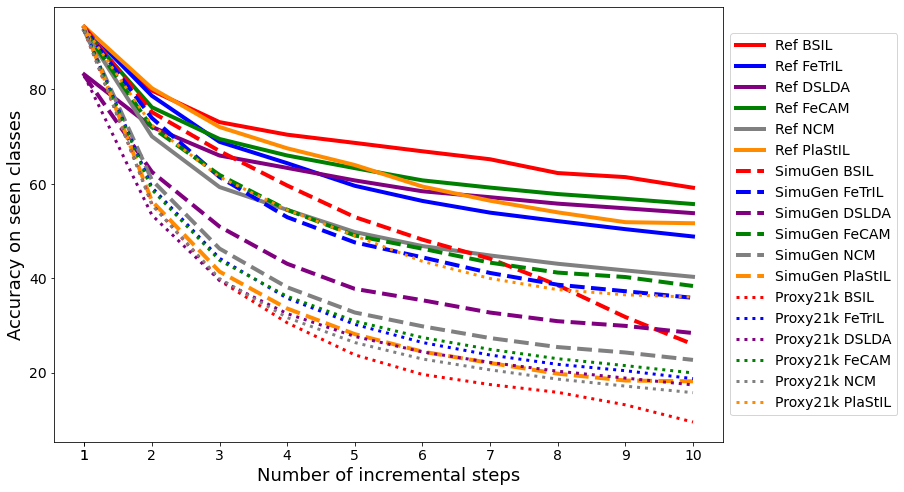}\label{fig_land}
      }
 \subfloat[iNat1k]{
      \includegraphics[height=0.18\textheight]{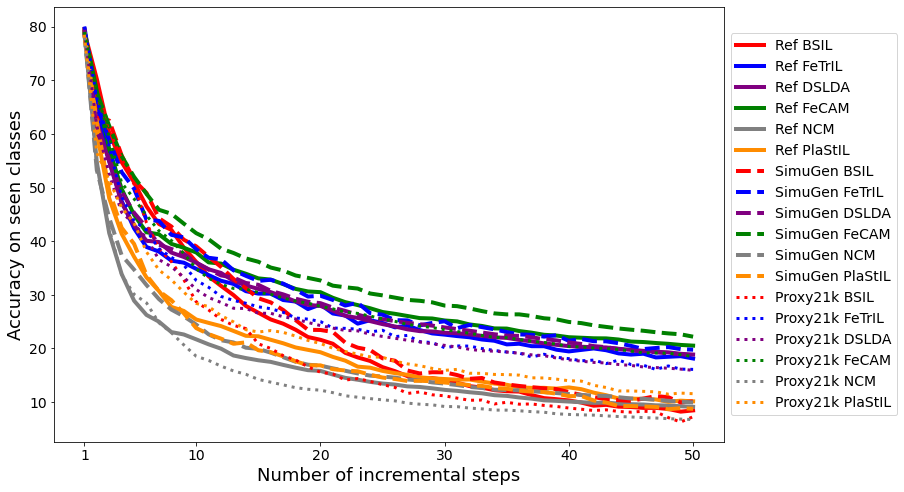} \label{fig_inat}
      }
\caption{Detailed incremental accuracy for (a) iNat1k with $\text{Card}(P_1)=20$ and $T=50$ steps and (b) Land1k with $\text{Card}(P_1) = 100$ and $T = 10$ steps, and their corresponding simulated datasets obtained following SimuGen and Proxy21k.}
\label{fig_learning_curves}
\end{figure}

The results presented in Table~\ref{tab_results_reco_agg} show that SimuGen and Proxy21k are effective since they recommend algorithms whose accuracy on the real datasets $\mathcal{D}$ are close or equal to that of the best algorithm (oracle).
In most settings, they are also better than the considered baselines. 
On average, the best scores are obtained with $\rho_{gen}$, whose recommendations are only 0.32 accuracy points below the oracle $\rho_{ref}$ when all simulation steps are run. 
Proxy21k also gives interesting results, but the gap with the oracle is higher than for SimuGen (1.20 points). 
A closer look at the results shows that Land1k is much better approximated by the data simulated with SimuGen compared with Proxy21k. This is further illustrated by the example of Figure~\ref{fig_land}, where SimuGen simulations are closer to the experiments with the reference datasets than the Proxy21k simulations.
The same is true for iNat1k (Figure~\ref{fig_inat}) despite this dataset being well covered by ImageNet-21k, the visual knowledge base used by Proxy21k.
Our incremental settings are closer to the same subset of AdvisIL experiments for which FeCAM performs better. Here, AdvisIL always recommends FeCAM because the precomputed experiments presented in \cite{feillet2023advisil} focus on incremental parameters with datasets of 100 classes and small memory budgets, advantaging fixed-representation algorithms over fine-tuning based algorithms. 
This highlights AdvisIL's lack of flexibility beyond its initial configuration. 
We conclude that simulation using generative models enables a better and more flexible approximation of the incremental data stream than using a preexisting visual knowledge base. 

Among the baselines that recommend a fixed algorithm, FeCAM has the best accuracy on average, followed by FeTrIL and DSLDA. 
These methods perform better when the feature extractor is trained on a larger data subset because the obtained deep representation is more transferable. 
FeCAM separates classes particularly well as it can handle the heterogeneity of scales for the different class distributions.
Figure~\ref{fig_learning_curves} shows that its initial accuracy is comparable to that of FeTrIL and BSIL while its accuracy is more stable.
DSLDA also has a stable performance across the incremental process but it is penalized by a lower initial accuracy due to its LDA classifier. 
The features of the initial classes are more difficult to separate with the LDA-based classifier compared with the linear layer of BSIL with which the representation has been optimized, and also the Bayesian classifier of FeCAM or the linear SVCs of FeTrIL.
BSIL performs better when the number of incremental steps is small. Despite a knowledge distillation loss and a rebalanced softmax loss to emulate past classes, this algorithm is penalized by the difficulty of preserving an adapted representation of past classes when the incremental sequence is long. 
While NCM is the simplest method, it still outperforms BSIL and PlaStIL in the challenging scenario where $Card(P_1)$=500 and T=101.
\begingroup
\setlength{\tabcolsep}{3pt} 
\renewcommand{\arraystretch}{1.05} 
\begin{table}
\begin{center} 
\caption{Performance gap between the average incremental accuracy of the methods proposed by the oracle ($\rho_{ref}$) and that of the algorithms recommended with different methods (results averaged over (i) the three datasets (ii) the six DFCIL settings of the form $(\text{card}(P_1), T)$, (iii) all settings). Individual results are in the supplementary. P : PlaStIL, B: BSIL, N:NCM, D:DSLDA, F:FeTrIL, Fc: FeCAM. Gaps closer to zero are better. Best results in bold, second best underlined. 
}  
\label{tab_results_reco_agg}
\resizebox{0.99\linewidth}{!}
{
\begin{tabular}{|l|l|l|lll|lll|l|llllll|} 
\hline
\multicolumn{3}{|l|}{}  & \multicolumn{13}{|l|}{ \color{white}Some more space in here\color{black}Accuracy gap for recommendation methods} \\ 
\cline{3-16}
\multicolumn{2}{|l|}{}  & $\rho_{ref}$  & $\Delta_{gen}^T$ & $\Delta_{gen}^{\text{effi}}$ & $\Delta_{gen}^3$ & $\Delta_{proxy}^T$ & $\Delta_{proxy}^{\text{effi}}$ & $\Delta_{proxy}^3$ & $\Delta_{adv}$ & $\Delta_P$ & $\Delta_B$ & $\Delta_N$  & $\Delta_D$ & $\Delta_F$ & $\Delta_{Fc}$ \\ \hline 
\multicolumn{1}{|l|}{\raisebox{-2.5\normalbaselineskip}[0pt][0pt]{\rotatebox[origin=c]{90}{Scenario}}} & 
                                                       (20, 49)   & 35.12  &  0.0  &  -3.95  &  -3.95  &  0.0  &  -2.55  &  -6.5  &  0.0  &  -11.68  &  -11.54  &  -17.0  &  -2.31  &  -4.35  &  0.0 \\ 
\multicolumn{1}{|l|}{}                               & (100, 10)  & 57.22  &  -1.41  &  0.0  &  0.0  &  -2.93  &  -1.66  &  -1.66  &  -4.11  &  -6.33  &  0.0  &  -12.88  &  -6.89  &  -5.26  &  -4.11  \\
\multicolumn{1}{|l|}{}                               & (200, 5)   & 66.94  &  0.0  &  0.0  &  0.0  &  -1.1  &  -1.1  &  -1.1  &  -7.3  &  -5.73  &  0.0  &  -11.44  &  -10.54  &  -7.12  &  -7.3 \\
\multicolumn{1}{|l|}{}                               & (500, 5)   & 71.06  &  -0.14  &  0.0  &  0.0  &  -2.64  &  -1.99  &  -1.86  &  -2.84  &  -6.05  &  0.0  &  -3.14  &  -6.18  &  -3.66  &  -2.84 \\
\multicolumn{1}{|l|}{}                               & (500, 11)  & 67.98  &  -0.26  &  -0.26  &  -3.5  &  -0.25  &  -0.22  &  -1.05  &  -0.05  &  -9.73  &  -3.45  &  -0.22  &  -3.21  &  -1.65  &  -0.05 \\
\multicolumn{1}{|l|}{}                               & (500, 101) & 67.8  &  -0.11  &  -0.16  &  -0.89  &  -0.27  &  -0.16  &  -0.16  &  -0.11  &  -60.25  &  -59.78  &  -0.16  &  -3.12  &  -3.63  &  -0.11 \\ \hline
\multicolumn{1}{|l|}{\raisebox{-1.1\normalbaselineskip}[0pt][0pt]{\rotatebox[origin=c]{90}{Dataset}}}      
                                                     & IN1k       & 49.07  &  -0.08  &  -1.98  &  -3.18  &  -0.71  &  -0.08  &  -1.98  &  -1.18  &  -14.2  &  -11.25  &  -5.09  &  -3.22  &  -3.0  &  -1.18 \\
\multicolumn{1}{|l|}{}                               & iNat1k     & 60.82  &  -0.07  &  -0.02  &  -0.44  &  -0.39  &  -1.27  &  -1.71  &  -3.01  &  -18.45  &  -12.21  &  -7.08  &  -6.58  &  -4.48  &  -3.01 \\ 
\multicolumn{1}{|l|}{}                               & Land1k     & 73.17  &  -0.81  &  -0.19  &  -0.55  &  -2.49  &  -2.49  &  -2.47  &  -3.02  &  -17.23  &  -13.93  &  -10.26  &  -6.32  &  -5.35  &  -3.02\\
\hline
\multicolumn{2}{|l|}{Average}   &  61.02  &  \textbf{-0.32}  &  \underline{-0.73}  &  -1.39  &  -1.2  &  -1.28  &  -2.06  &  -2.4  &  -16.63  &  -12.46  &  -7.47  &  -5.37  &  -4.28  &  -2.4 \\ 
\hline
\end{tabular}
}
\end{center}
\end{table}
\endgroup
In this scenario, it benefits from a highly transferable feature extractor.
Although on par with BSIL and DSLDA on short scenarios, PlaStIL has the lowest average accuracy across all experiments. 
Its partial fine-tuning mechanism is pushed to its limits by long sequences. 
On average, the most challenging scenario for all methods is the one with 50 steps of 20 classes each (Figure \ref{fig_inat}). 
The representation learned from the 20 initial classes hardly generalizes to the rest of the stream, and fine-tuning struggles with the numerous small incremental steps. \\
\quad We provide additional results in the supplementary material, showing the relevance of SimuGen for recommending the second-best algorithm and determining which algorithm to discard in priority. 
SimuGen recommendations are also more stable than those of Proxy21k when considering only $c$ candidate algorithms out of the six, with $c \in [2, 5]$.

\subsection{Analysis of recommendation dynamics}
\label{subsec_dynamics}
\begin{figure*} 
 \centering 
 \subfloat[(20,50)]{%
      \includegraphics[height=0.094\textheight]{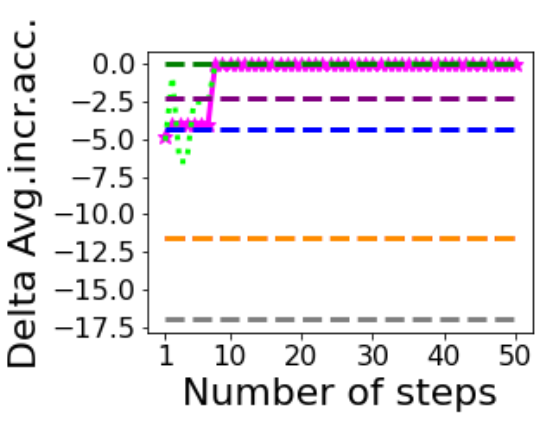}}  
      \label{fig_B20T50}  \hspace{0.1em}
 \subfloat[(100,10)]{%
      \includegraphics[height=0.094\textheight]{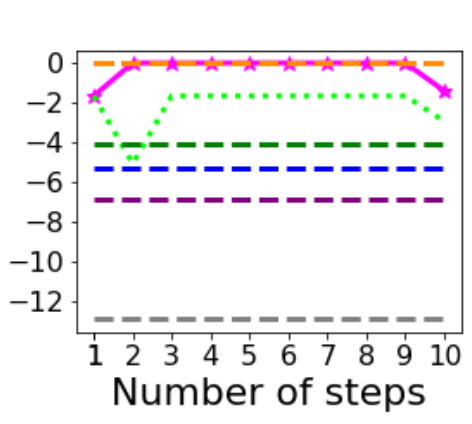}}     \hspace{0.1em}
 \subfloat[(200,5)]{%
      \includegraphics[height=0.094\textheight]{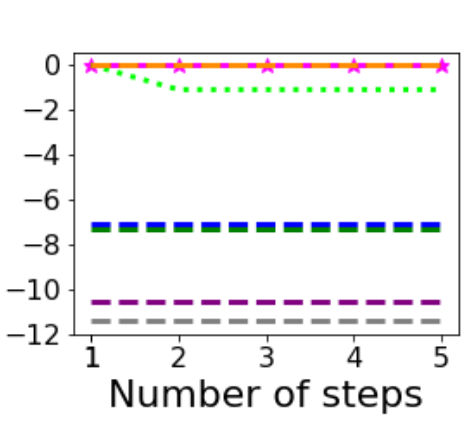}}     \hspace{0.1em}
 \subfloat[(500,11)]{%
      \includegraphics[height=0.094\textheight]{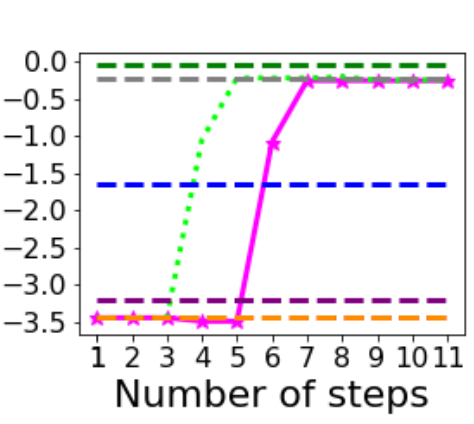}}     \hspace{0.1em}
 \subfloat[(500,101)]{%
      \includegraphics[height=0.094\textheight]{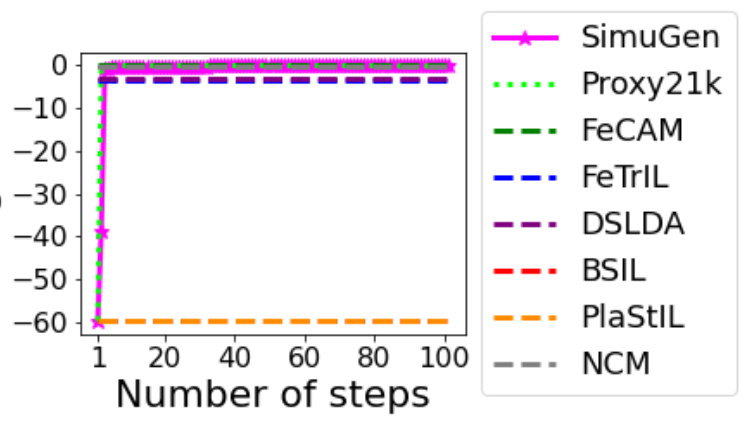}}    
 \caption{Detailed view of the performance gap between the oracle and the recommendation methods after simulating $t = 1, 2, \dots T$ incremental steps, for scenarios of the form ($Card(P_1)$, T). Results are averaged over the three datasets (IN1K, iNat1k, Land1k).}
  \label{fig_dynamics}
\end{figure*}

So far, we have computed recommendations on the simulated datasets by computing the average incremental accuracy of each algorithm after the $T$ steps of the incremental process. 
In Figure \ref{fig_dynamics}, we show how the relevance of recommendations evolves with the number of incremental learning steps performed on the simulated datasets. 
For an incremental setting $(\text{Card}(P_1), T)$, we show the accuracy gap between the oracle $\rho_{ref}$ and the recommendations made after performing $t$ incremental learning steps with $t \leq T$.
In Table~\ref{tab_results_reco_agg}, we see that the more efficient strategies consisting in (i) running only three steps or (ii) running three steps then pruning the set of candidate algorithms perform better than the choice of any fixed algorithm. 
Except for the case of $(\text{Card}(P_1)=100, T=10)$ where BSIL is the best algorithm for all three datasets starting from the initial step, Figure \ref{fig_dynamics} shows the interest of both Proxy21k and SimuGen simulation approaches, even if the DFCIL experiments are only partially executed. We provide more detailed plots in the supplementary material. 

%% file: sec/7_discussion.tex
\subsection{Discussion}
\textbf{Performance.} In this article, we have introduced a method for recommending a relevant DFCIL algorithm for applying it to a future data stream. 
We proposed two approaches for building a simulated data stream from the same visual domain as the initial dataset, one using generative models (SimuGen) and one using existing databases (Proxy21k). 
An evaluation with six challenging incremental scenarios and three large-scale datasets covering various visual domains shows that SimuGen provides the best results.
Proxy21k also outperforms fixed algorithms, but is less flexible than SimuGen. 
The overall performance of both approaches outperforms the fixed choice of any of the DFCIL algorithm we tested. 
In addition to its usefulness for recommending DFCIL algorithms, this work constitutes a stress test for the algorithms studied. It shows that none is the best in all scenarios, and therefore underlines the importance of a comprehensive evaluation of CL algorithms to understand their strengths and limitations.

\textbf{Novelty.} To the best of our knowledge, this method is the first to simulate an entire incremental learning process using a generative model and to consider the semantic content of the incremental dataset to recommend a relevant DFCIL algorithm. 
Our method facilitates the deployment of continual learning. 
It requires little domain-related expertise from the user, as the user only needs to provide a few DFCIL settings and an initial data subset. 
The method leverages the knowledge encoded in a pretrained LLM or a knowledge base to simulate a data stream from the same domain as the input data.

\textbf{Further applications.} 
We have chosen the data-free (DFCIL) case because it is challenging and covers the deployment of continual learning in resource-constrained environments, such as embedded systems~\cite{hayes2022embedded,verwimp2023applications}.
We did not include algorithms relying on a memory buffer, as the comparison with DFCIL algorithms would be unfair~\cite{belouadah2021_study,masana2022survey}. 
However, the recommendation method can be applied to other continual learning scenarios defined by different data stream structures~\cite{van2022three}, provided that the working hypotheses described in Subsection~\ref{subsec_hyp} are satisfied. 
For example, it can be applied to CIL with a memory buffer or task-incremental learning.
It can also be adapted for domain-incremental learning, in which the set of classes is fixed, but the distribution of the classes changes.

We discuss below the limitations of our method, as well as potential mitigation measures. 

\textbf{Cost of recommendation.}
A preselection of candidate algorithms can readily be applied considering practical criteria such as the possibility of updating the model on-device, the latency of a model update, or the required storage. 
Such preselection would lower the cost induced by running multiple candidate algorithms.
Additionally, training low-performing algorithms can be stopped early, as we propose to do with the ``explore then prune'' strategy.
We also show in Figure \ref{fig_dynamics} that in most settings, running half of the simulation steps is enough to get an accurate recommendation because algorithm rankings are stable at the end of the incremental process.

\textbf{Cost of data generation.} 
The performance improvement brought by the recommendations with SimuGen comes at an initial computational cost due to the use of generative models. 
Following usual practice in continual learning~\cite{goswami2024fecam,hayes2020_deepslda,janson2022simple, petit2024analysis}, we consider that the first step is offline and non-incremental. 
Consequently, using large generative models as a preliminary is not a strong limitation in the considered setting.
The cost of data generation could be reduced using more efficient textual~\cite{xiao2023smoothquant} and visual generation models~\cite{li2024snapfusion}. 
Another option would be to use Proxy21k when the initial dataset is well covered by an existing visual dataset and SimuGen when this is not the case. 
A similarity threshold between the user's data subset and an existing dataset could be used to select one of the two methods.

\textbf{Relevance of simulated data.}
We observed that the proposed new class names sometimes lacked diversity from one prompt to another, hence the multiple runs with diversified prompts. 
To improve the outputs of the LLM, an automatic cleaning process that limits hallucinations or peculiar outputs highlighted in~\cite{ye2023cognitive} could be used.
Another way of cleaning the data could consist in checking the existence of the proposed class names against a knowledge base such as Wikidata. 
Also, the LLM does not always follow the explicit instructions from the prompt. 
For instance, a valid JSON output is not always provided despite an explicit requirement.
This behavior has a minor effect on the simulation results but complicates data processing.
Moreover, pretrained text-to-image models might offer limited coverage of the targeted visual domain. 
Ensuring visual diversity and semantic coherence within an image set obtained from the same textual prompt but with different random seeds also remains an open problem~\cite{grimal2024tiam}.
However, generation models are trained with increasingly large and diversified datasets, and this will increase their usability for simulating incremental processes. 

%% file: sec/8_conclusion.tex
\section{Conclusion}
\vspace{-0.6em}
Despite intensive research in the field, no existing algorithm works best in all DFCIL scenarios~\cite{belouadah2021_study,feillet2023advisil}. 
This makes algorithm recommendation necessary for optimal deployment. 
Our work is the first to recommend an incremental learning algorithm based on a simulated data stream for a user-defined scenario.
This is a direction of research that is still relatively unexplored, but is useful for the practical adoption of CIL~\cite{feillet2023advisil}. 
We show that by leveraging generative models, we can accurately recommend DFCIL algorithms in various visual domains and incremental scenarios. 
We plan to study the impact of the data stream structure on the performance of DFCIL algorithms, experiment with other continual learning scenarios, and investigate further possible usage of generative models to improve incremental learning.

%% file: sec_supp/0_ref_datasets.tex
\section{Reference datasets and metadata}
\label{sec:datasets}

In our experiments, from the reference datasets ILSVRC\footnote{\url{https://www.image-net.org/index.php}}~\cite{olga2015_ilsvrc}, iNaturalist2018\footnote{\url{https://github.com/visipedia/inat_comp/blob/master/2018/}} \cite{van2018inaturalist} and Google Landmarks v2\footnote{\url{https://github.com/cvdfoundation/google-landmark}}~\cite{noh2017_landmarks}, we form three balanced thousand-class-datasets denoted IN1k, iNat1k, and Land1k respectively. 
IN1k corresponds to a generic fine-grained classification task. iNat1k covers fine-grained fauna, flora and fungi species. Land1k contains pictures of various landmarks from all around the world.

We obtain class descriptions either from a knowledge base such as WordNet, when available, or by prompting an LLM. 
Here most of the class names of the three datasets can be linked to a metadata source. 
Each class from ILSVRC corresponds to a WordNet synset; each class from iNaturalist2018 corresponds to a living species and is provided with the species' ancestors in a natural taxonomy; most Google Landmarks v2 images are linked to their original Wikipedia pages, and landmark categories are provided. This way, most of the class names can be mapped to a WordNet synset.
When a definition if available in WordNet for this synset, we take it as class description. Otherwise, we prompt an LLM to obtain it (see example with Land1k in Section \ref{sec:text}). 
So in all cases, we are able to enrich our prompts when applying SimuGen approach to IN1k, iNat1k and Land1k.

Next, we use the class names and their description to prompt an LLM and obtain new class names and descriptions. 

%% file: sec_supp/1_text.tex
\section{Simulation of class names} \label{sec:text}
The first step in our two dataset simulation approaches, SimuGen and Proxy21k, consists in obtaining a set of new class names. 

\subsection{SimuGen: Generation of new class names with Llama-2-13b-chat}
\label{subsec:comp_class_names}
In our implementation, we used the huggingface libraries \texttt{diffusers} (version 0.21.4) and \texttt{transformers} (version 4.34.1) along with the pretrained checkpoints of Llama-2-13b-chat\footnote{\url{https://github.com/facebookresearch/llama}} \cite{touvron2023llama}.

For each of the three datasets (IN1k, iNat1k, Land1k), following the SimuGen approach, we prompt Llama-2-13b-chat with pairs that are each composed of a class name and a description of the class name. 
We use NLTK version 3.8.1 to query the WordNet corpus for synsets and definitions. 

In the following, we provide a few examples of prompts and outputs.
We observe that even when example descriptions are short or not very specific, the output is satisfying both in terms of length and content.

\textbf{Example 1. IN1k} 
For IN1k, we form the set of class names by taking the WordNet synset corresponding to each class identifier. We take the WordNet definition of the synset as the description of the class. 

(a) Input 

\begin{lstlisting}[language=json,firstnumber=1]
{
{
'role': 'system', 
'content': 'Always answer in JSON format'
}, 
{
'role': 'user', 
'content': 'Here is a JSON containing concept names with their description: 
{
"concept": [ 
{
"name": "restaurant", 
"description": "a building where people go to eat"
}, 
{"name": "violin", 
"description": "bowed stringed instrument that is the highest member of the violin family;"
}, 
{"name": "chain mail", "description": "(Middle Ages) flexible armor made of interlinked metal rings"
},
] 
}
Could you provide 10 more concept names, with a short visual description of each item?'
}
}
\end{lstlisting}

(b) Output

\begin{lstlisting}[language=json,firstnumber=1]
Sure, here are ten more concept names with short visual descriptions in JSON format:

{
"concept": [
{
"name": "space shuttle",
"description": "a reusable spacecraft with a winged spaceplane and a rocket"
},
{
"name": "piano",
"description": "a musical instrument with a keyboard, pedals, and strings"
},
{
"name": "submarine",
"description": "a underwater vessel with a watertight hull and specialized equipment for navigation and communication"
}, ...
{
"name": "fountain pen",
"description": "a pen with a refillable ink reservoir and a nib"
}
]
}
\end{lstlisting}

\textbf{Example 2. iNat1k}

For iNat1k, each class corresponds to a natural species, i.e for an image whose file path is of the form ``05725$\_$Fungi$\_$Basidiomycota$\_$Tremellomycetes$\_$Tremel lales$\_$Naemateliaceae$\_$Naematelia$\_$aurantia/hash.jpg'', the species name is ``Naematelia aurantia''. The classes of iNaturalist-2018 can be mapped to 7 hierarchical levels in the natural taxonomy\footnote{\url{https://en.wikipedia.org/wiki/Taxonomy_(biology)}} (from fine to coarse : species, genus, family, order, class, phylum and kingdom). For each class, we select as a hypernym the lowest common ancestor of the class name (species) present in WordNet. If the species itself is present in WordNet, then we use the corresponding WordNet definition as the description of the class. Otherwise, we take the WordNet definition of the class hypernym as description. \\

(a) Input

\begin{lstlisting}[language=json,firstnumber=1]
[
{
'role': 'system', 
'content': 'Always answer in JSON format'
}, 
{
'role': 'user', 
'content': 'Here is a JSON containing fauna, flora and fungi species names with their description: 
    {
    "species": 
    [ {"name": "Gluphisia septentrionis", 
    "description": "moths and butterflies"}, 
    {"name": "Calendula officinalis", "description": "any of numerous chiefly annual herbs of the genus Calendula widely cultivated"}, 
    {"name": "Anacridium aegyptium", "description": "short-horned grasshoppers; true locusts"},
    ] 
Could you provide 10 more fauna, flora and fungi species names, with a short visual description of each item?'}
}
]

\end{lstlisting}

(b) Output

\begin{lstlisting}[language=json,firstnumber=1]
Sure, here are ten more fauna, flora, and fungi species names with short visual descriptions:

{"species": [
{"name": "Canis latrans", "description": "also known as the coyote, is a canid native to North America"},
{"name": "Lythrum salicaria", "description": "purple flowering plant found in wetlands and waterways"},
{"name": "Amanita muscaria", "description": "brightly colored and poisonous fungi with distinctive cap and stem"},
...
{"name": "Cygnus olor", "description": "large, long-necked bird with a distinctive white plumage"}]}
\end{lstlisting}

\textbf{Example 3. Land1k}
For Land1k, we use the metadata\footnote{\url{https://s3.amazonaws.com/google-landmark/metadata/train_label_to_hierarchical.csv}} provided along with the images to determine the class names (``category'') and their hypernyms (``hierarchical$\_$label''). 
For the classes whose hypernym matches a WordNet synset, we take the WordNet definition of the synset as the class description.
For the classes without available metadata, we use Llama-2-13b-chat to get descriptions. 
In this case, similarly to the above examples, we use a system prompt asking for an output in JSON format, and ask for descriptions using the following prompt: \texttt{``Here is a list of landmarks <list>. Can you provide a short visual description of each item ?''}. 

(a) Input 

\begin{lstlisting}[language=json,firstnumber=1]
[
{
'role': 'system', 
'content': 'Always answer in JSON format'
}, 
{
'role': 'user', 
'content': 'Here is a JSON containing landmark names with their description: 
{"landmark": [ {"name": "Pester", "description": "Pester, mountain, mountain"}, 
{"name": "Niagara Falls", "description": "Niagara Falls, waterfall, waterfall"}, 
{"name": "Khotyn Fortress", "description": "Khotyn Fortress, castle / fort, fortress"},
] 
Could you provide 10 more landmark names, with a short visual description of each item?'}
}
]
\end{lstlisting}

(b) Output

\begin{lstlisting}[language=json,firstnumber=1]
Sure, here are ten more landmark names with short visual descriptions in JSON format:
{
"landmark": [
{
"name": "Eiffel Tower",
"description": "Eiffel Tower, Paris, France - Iron lattice tower with observation decks"
},
{
"name": "Great Barrier Reef",
"description": "Great Barrier Reef, Australia - Largest coral reef system in the world"
},
{
"name": "Machu Picchu",
"description": "Machu Picchu, Peru - Inca citadel built on a mountain ridge"
},
...
{
"name": "Petra",
"description": "Petra, Jordan - Ancient city carved into sandstone cliffs"
}]}
\end{lstlisting}

\subsection{Proxy21k: Selection of class names using WordNet}

Following our approach Proxy21k, we randomly sample a thousand leaf classes from the ImageNet subtree related to the visual domain of each dataset. In our implementation, we use the WordNet package (version 1:3.0-37) for obtaining the synsets from a given visual domain. We define a visual domain by a set of keywords and run the following command with each keyword:  
\texttt{wn <keyword> -o -treen}. 
Then, we concatenate the outputs and parse the obtained lists to get the WordNet identifiers of the synsets. We obtain a list of ImageNet classes related to the visual domain of interest.

In the case of IN1k, we sampled class names from the complete ImageNet21k dataset.

In the case of iNat1k, we selected classes subsumed by the concepts ``flora'', ``fauna'' and ``fungi''.

In the case of Land1k, the WordNet synset ``landmark'' does not include enough classes relevant to the content of Land1k (and refers among others to craniometric points), so instead we selected class names from the subtrees defined by the hierarchical labels provided in the Google Landmarks v2 metadata.
This is the list of the 69 hierarchical labels from Google Landmarks v2 which can be found in WordNet: \textit{monument, park, mountain, road, museum, market, theater, church, castle, volcano, cemetery, ruin, temple, house, island, library, festival, lake, lighthouse, fountain, bridge, zoo, tower, mosque, city, prison, shopping, palace, cave, memorial, restaurant, factory, monastery, skyscraper, waterfall, river, trail, synagogue, settlement, observatory, government building, dam, hotel, square, harbor, canal, tree, garden, gate, stone, sculpture, mine, aqueduct, beach, school, artwork, air transportation, wetland, village, windmill, hospital, bath, stairs, stair, power plant, swimming pool, cross, pyramid, cliff}. 

\subsection{Comparison of class names}
In what follows, we provide additional illustrations for comparing the class names from IN1k, iNat1k, and Land1k and their simulated counterparts obtained with SimuGen and Proxy21k, respectively.

In Table~\ref{tab:min_dist}, for each dataset (IN1k, iNat1k, Land1k), we compute the percentage of new class names, obtained either with Llama-2-13b-chat (SimuGen) or WordNet (Proxy21k), whose nearest neighbor in the set of real class names is inferior or equal to a given value $d$ in terms of cosine distance. Class names are represented by their CLIP embeddings~\cite{radford2021clip}. 
As discussed in the article, the class names obtained following SimuGen tend to be closer to class names of IN1k, iNat1k and Land1k than the class names obtained following ProxyIN1k. In the case of IN1k however, the two methods yield very similar results. This is expected as class names from IN1k all come from WordNet. 
\begin{table}[ht] \centering
\caption{Percentage of simulated class names whose nearest neighbor among the real class names is inferior or equal to $d$. Class names are represented using their CLIP embeddings. Distances are expressed in terms of cosine distance. Best results in bold. A greater proportion of class names obtained with Llama-2-13b-chat is closer to real names.}
\begin{tabular}{|l|l|l|l|}
\hline
Dataset          & $d = 0.4$ & $d=0.2$ & $d=0.05$ \\ \hline
IN1k - WordNet   & 74.2    & 14.0  & 6.4    \\
IN1k - Llama     & \textbf{82.5}    & \textbf{16.4}  & \textbf{8.8}    \\ \hline
iNat1k - WordNet & 29.0    & 1.0   & 0.0    \\
iNat1k - Llama   & \textbf{86.7}   & \textbf{30.7}  & \textbf{5.1}    \\ \hline
Land1k - WordNet & 30.0    & 0.0   & 0.0    \\
Land1k - Llama   & \textbf{79.1}    & \textbf{24.3}  & \textbf{2.6}    \\ \hline
\end{tabular}
\label{tab:min_dist}
\end{table}

\begin{figure*}[ht]
    \centering
    \includegraphics[width=1\linewidth]{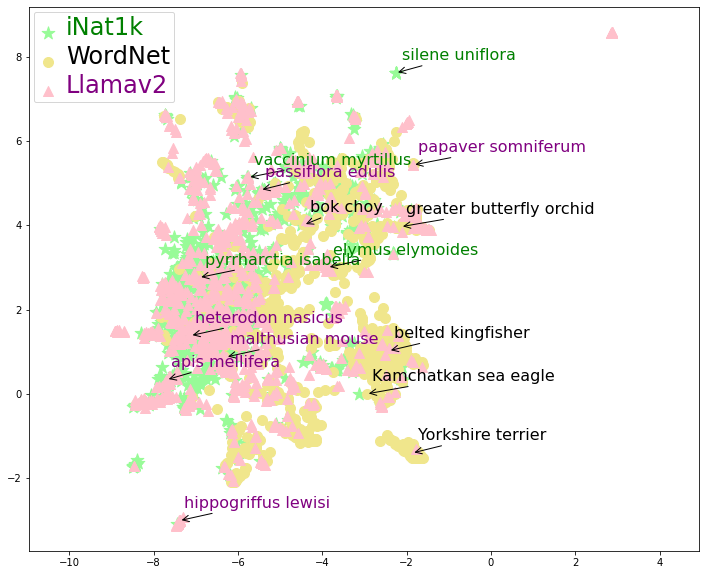}
    \caption{UMAP projection of the CLIP embeddings of the class names of iNat1k dataset where ``iNat1k'' denotes the original class names, ``WordNet'' the new class names obtained with Proxy21k and ``Llamav2'' those obtained with SimuGen.}
    \label{fig:umap_inat}
\end{figure*}

\begin{figure*}
    \centering
    \includegraphics[width=1\linewidth]{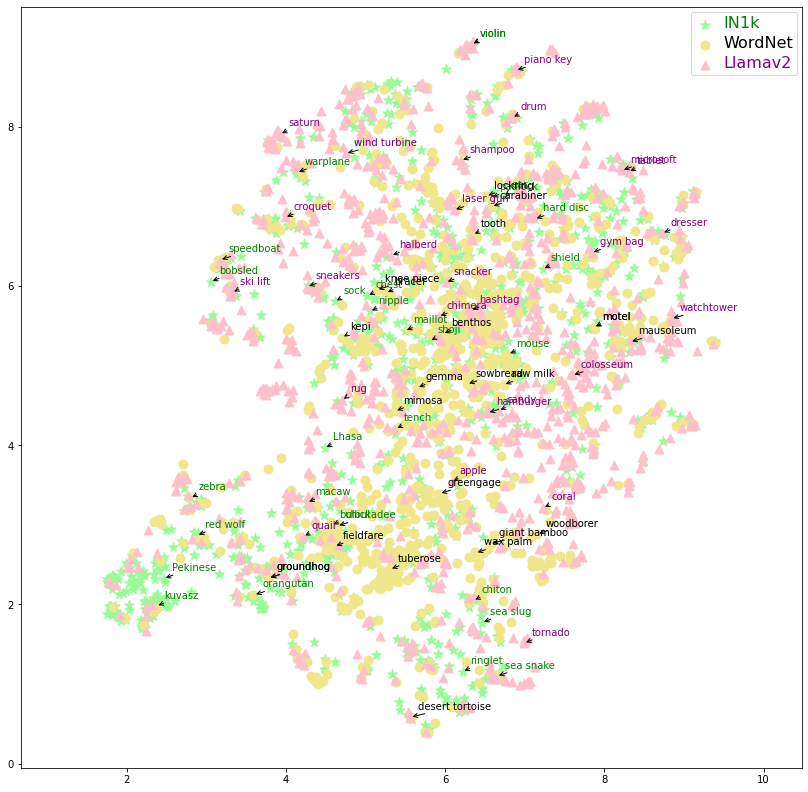}
    \caption{UMAP projection of the CLIP embeddings of the class names of IN1k dataset where ``IN1k'' denotes the original class names, ``WordNet'' the new class names obtained with Proxy21k and ``Llamav2'' those obtained with SimuGen.}
    \label{fig:umap_ilsvrc}
\end{figure*}

\begin{figure*}
    \centering
    \includegraphics[width=1\linewidth]{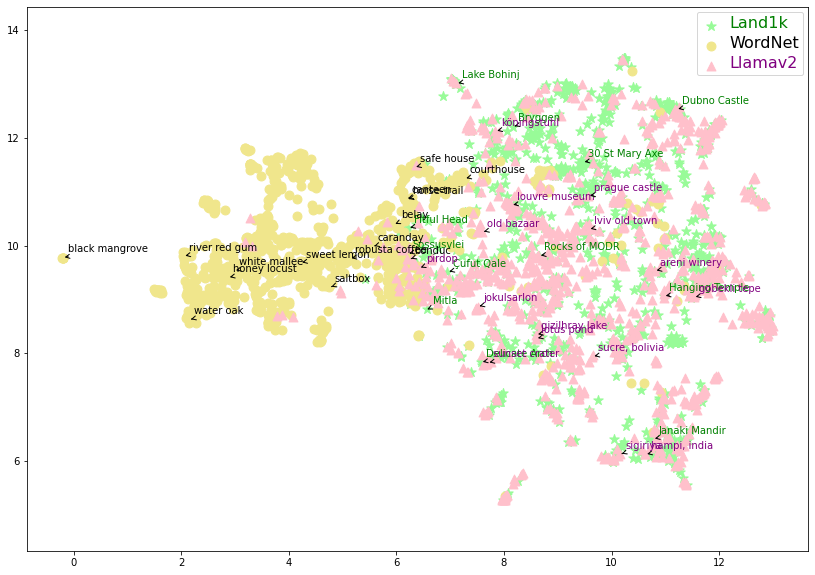}
    \caption{UMAP projection of the CLIP embeddings of the class names of Land1k dataset where ``Land1k'' denotes the original class names, ``WordNet'' the new class names obtained with Proxy21k and ``Llamav2'' those obtained with SimuGen.}
    \label{fig:umap_land}
\end{figure*}

We show the UMAP projections of the CLIP embeddings of class names for iNat1k, IN1k and Land1k in Figure~\ref{fig:umap_inat}, Figure~\ref{fig:umap_ilsvrc} and Figure~\ref{fig:umap_land}, respectively. A quarter of the class names from each dataset is randomly selected for the visaulization. The overlap of points associated with SimuGen and iNat1k class names is high, while it is lower for Proxy21k and iNat1k.  

In the case of IN1k (Figure \ref{fig:umap_ilsvrc}), the class names obtained from the three different ways are quite homogeneously spread and tend to overlap, as expected. \\
In the case of iNat1k (Figure \ref{fig:umap_inat}), the overlap of points associated with SimuGen (Llamav2) and iNat1k class names is high, while it is lower for Proxy21k (WordNet) and iNat1k. The LLM respects the implicit constraint of providing class names in Latin, whereas WordNet contains many species names in their vernacular form, resulting in higher similarity. 
Although Llamav2 is better at generating class names similar to those provided as examples in the prompts, limitations are highlighted in Figure \ref{fig:umap_inat}. For instance, ``hippogriffus lewisi" is the Latinized name of a mythical animal, and ``Malthusian mouse" is a figurative expression and not an actual natural species. We have chosen not to filter these classes, as there are not many of them. \\
The strongest difference between the class names obtained following SimuGen and following Proxy21k is observed for Land1k. In Figure \ref{fig:umap_land}, we see that class names obtained following Proxy21k (WordNet) are almost separated from the others. This is partly due to the hierarchical label ``tree" from the Google Landmarks v2 metadata. This synset ``tree" is responsible for numerous class names corresponding to \textit{tree species} instead of \textit{landmarks} associated to remarkable trees.  
Additionally, we observe that the classes selected by the Proxy21k approach are quite general e.g. ``cavern'', ``clock tower'' and ``flea market''. Although WordNet contains numerous examples of landmarks, the database rarely contains enough samples of the same landmark to populate a class for our DFCIL experiments. 
In contrast, the class names obtained with Llama-2-13b-chat are more specific, making them closer to the actual content of Land1k, e.g. ``Luray Caverns'', ``Izmir Clock Tower'' or ``Monastiraki Flea Market''.

Finally, we examine the intersection between the sets of new class names obtained with our two simulation approaches. 
This intersection includes less than $5\%$ of common class names on average for the three datasets. 
Around $20\%$ of class names can be matched approximately using substring matching to associate e.g. ``Alaskan brown bear'' with ``brown bear'', ``death cap'' with ``death cap mushroom'', or ``crystal'' with ``crystal chandelier''. 
The low intersection can be explained by the fact that the visual domains of the three datasets is very wide, and the same visual concepts can be described by various textual labels, let alone the variations between vernacular names and scientific names in the case of iNat1k. 

%% file: sec_supp/2_image.tex
\newpage
\section{Simulation of images}

\subsection{Generation of images with stable-diffusion-2-1-base}

Our implementation of image generation is based on the huggingface libraries \texttt{diffusers} (version 0.21.4) and \texttt{transformers} (version 4.34.1) and uses the pretrained checkpoints of stable-diffusion-2-1-base\footnote{\url{https://huggingface.co/stabilityai/stable-diffusion-2-1-base}}.
We use the following hyperparameters for the generative model:
\begin{itemize}
    \item[$-$] float precision: float16,
    \item[$-$] random seed: from zero to $n$ where $n$ is the number of images per class,
    \item[$-$] guidance scale: 2.0 ,
    \item[$-$] number of inference steps: 50 (default),
    \item[$-$] image size: 512x512 (default),
    \item[$-$] scheduler for denoising: PNDM (default),
    \item[$-$] text encoder and tokenizer for prompts: CLIP (default).
\end{itemize}
The guidance scale value recommended by the authors of \cite{stablediff} is around 7.0. 
Lower values result in lower image quality but in higher visual diversity. 
Our choice of guidance scale (2.0) follows the protocol reported in \cite{sariyildiz2023fake}: since we generate images with the purpose of learning from them, not displaying them, we choose to trade a bit of visual quality against more visual diversity.

\begin{figure*}[!ht] 
 \centering
 \subfloat[]{%
      \includegraphics[width=0.26\textwidth]{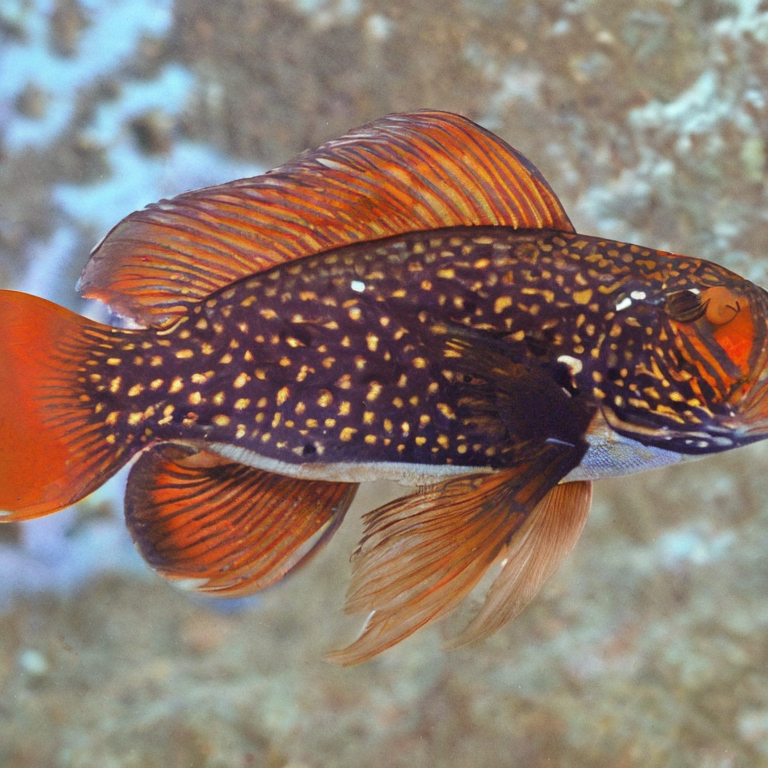} \label{fig:cod1}
      } \qquad
 \subfloat[]{
      \includegraphics[width=0.26\textwidth]{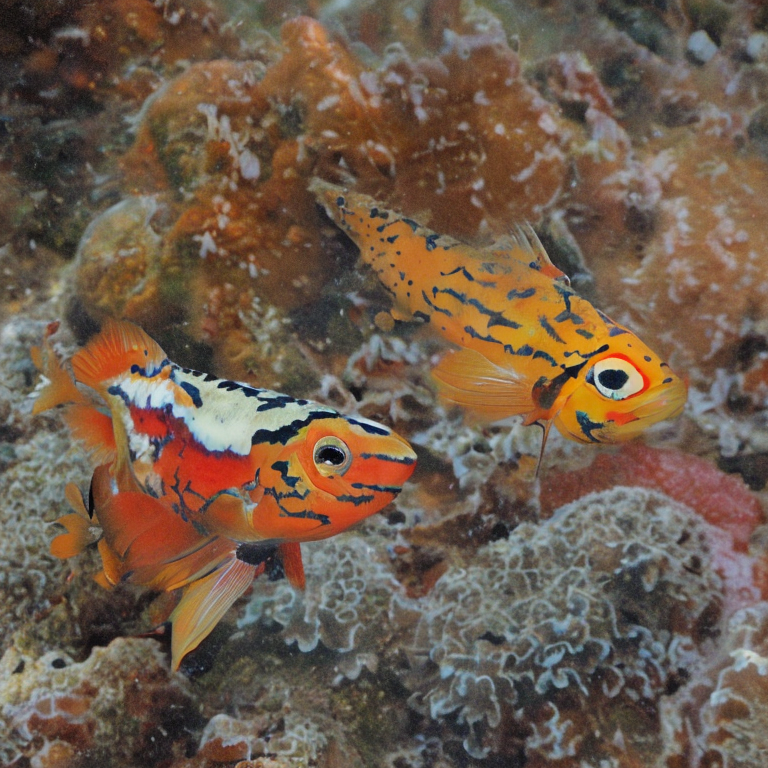}   \label{fig:cod2}
      } \qquad
 \subfloat[]{
      \includegraphics[width=0.26\textwidth]{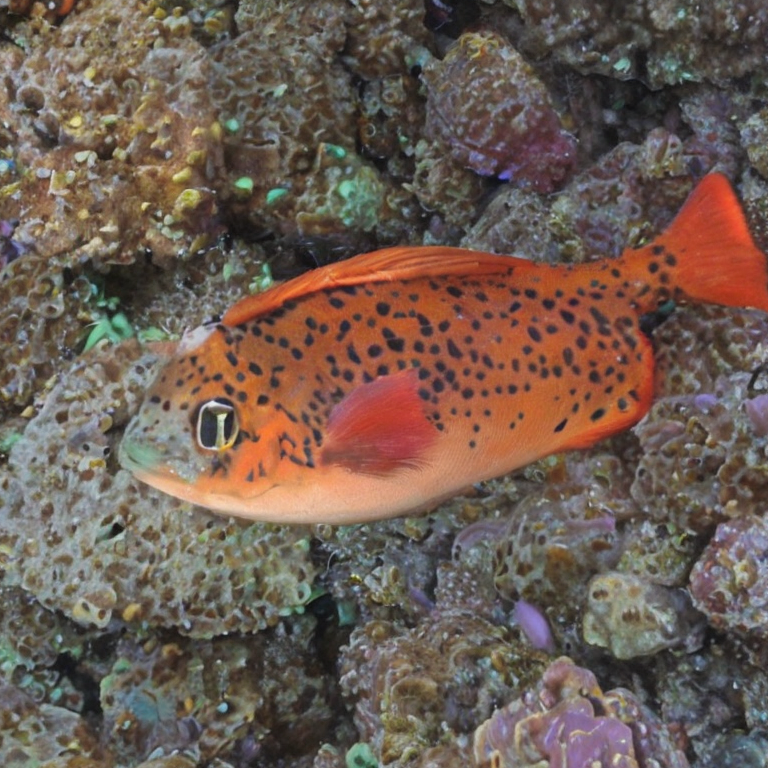} \label{fig:cod3}
      } \qquad
 \caption{Example images for the prompt ``a photo of a Lycodes pyrioides, a species of cod found in the north Pacific Ocean, characterized by its vibrant orange color and distinctive markings''.}%
 \label{fig:cod}
\end{figure*}

\begin{figure*}[!ht] 
 \centering
 \subfloat[]{%
      \includegraphics[width=0.26\textwidth]{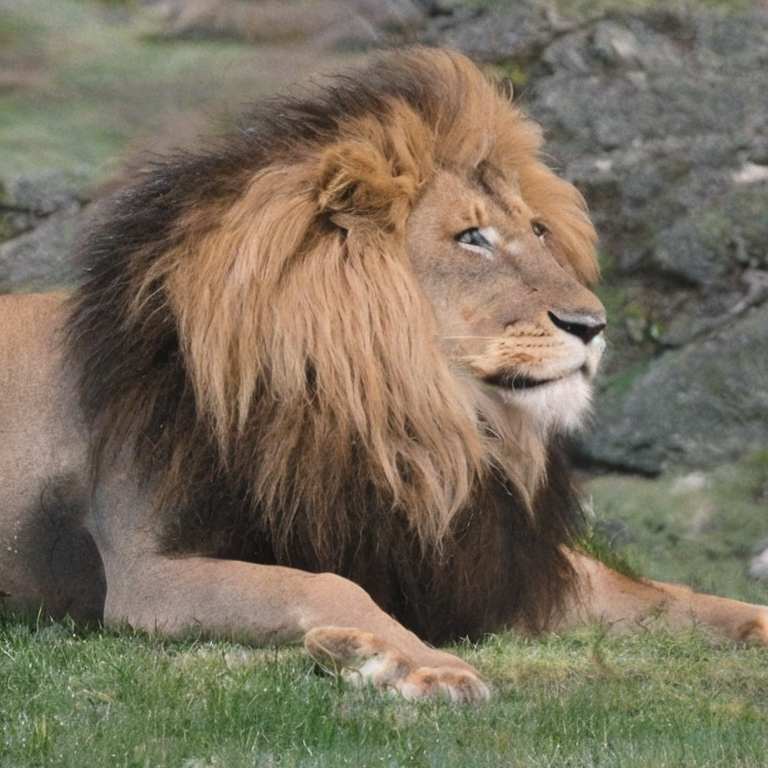}
      } \qquad
 \subfloat[]{
      \includegraphics[width=0.26\textwidth]{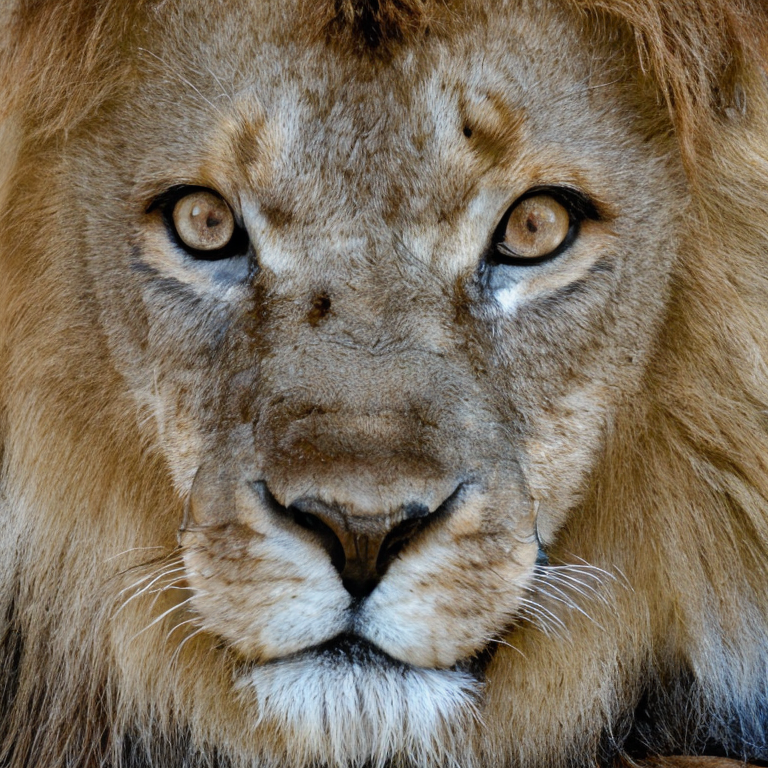} 
      } \qquad
 \subfloat[]{
      \includegraphics[width=0.26\textwidth]{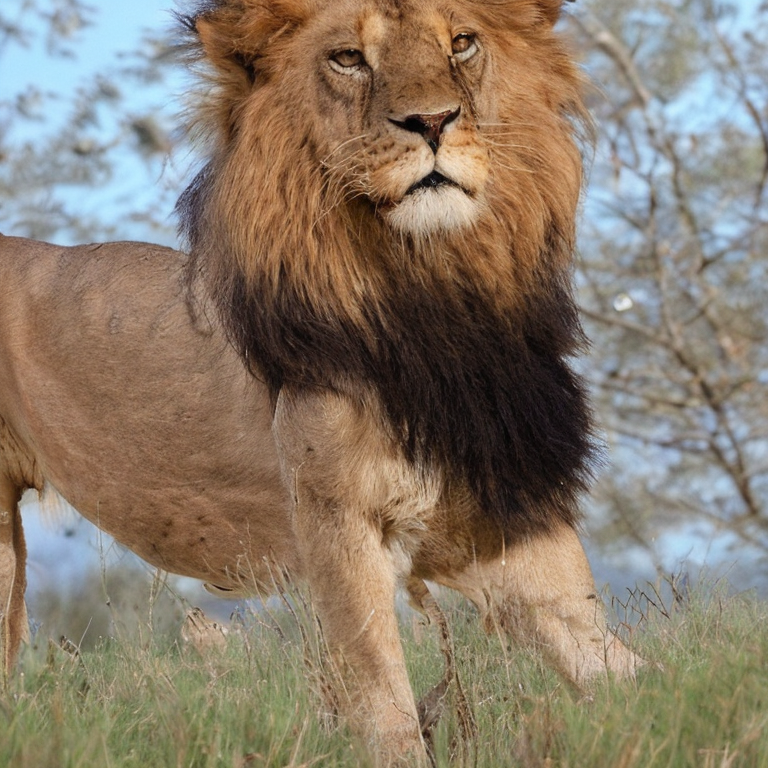}
      } \qquad
 \caption{Example images for the prompt ``a photo of a Lion, a large carnivorous mammal with a short, tawny or golden coat and a long mane''.}%
 \label{fig:lion}
\end{figure*}

\begin{figure*}[!ht] 
 \centering
 \subfloat[]{%
      \includegraphics[width=0.26\textwidth]{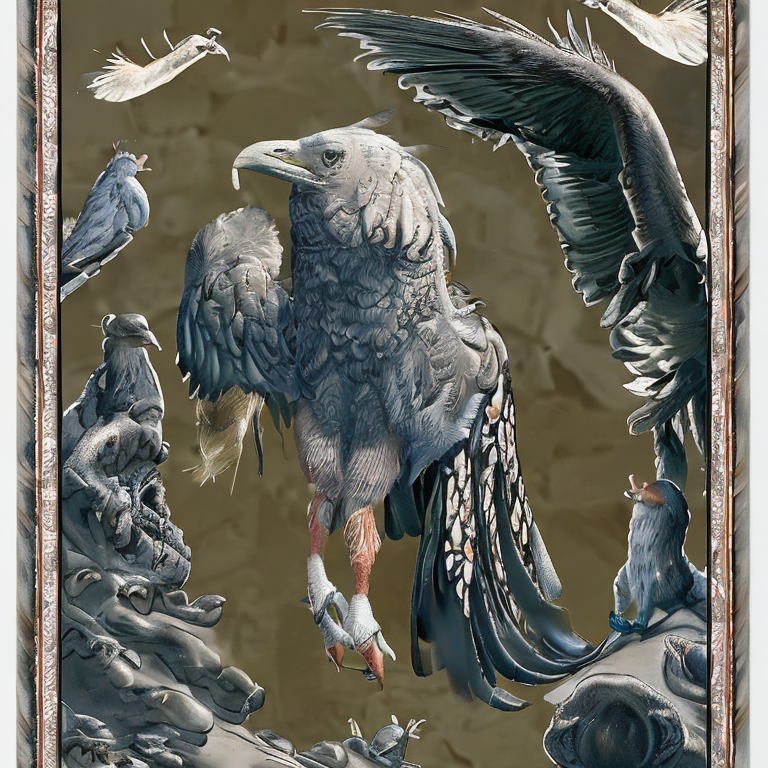} 
      } \qquad
 \subfloat[]{
      \includegraphics[width=0.26\textwidth]{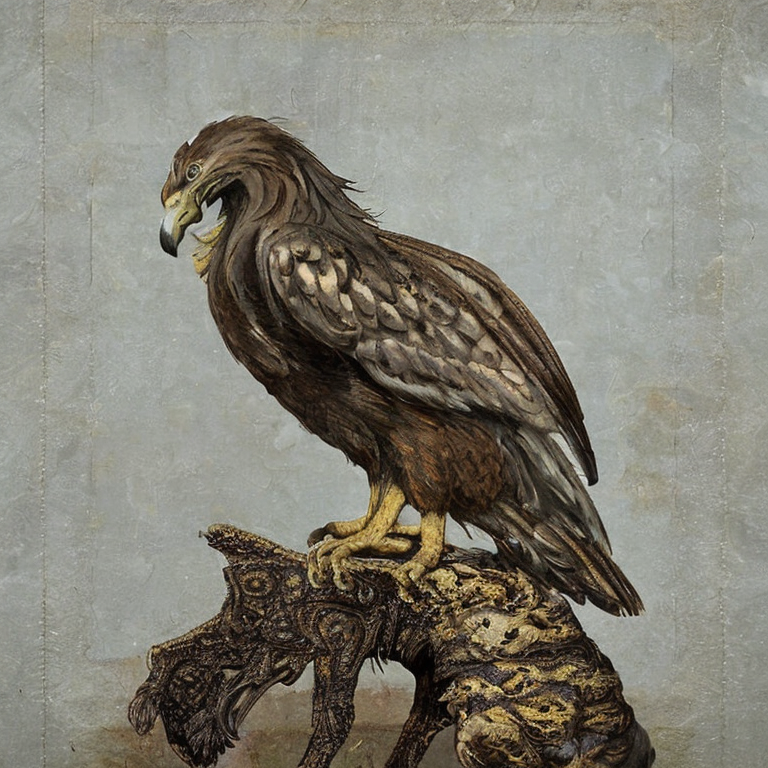} 
      } \qquad
 \subfloat[]{
      \includegraphics[width=0.26\textwidth]{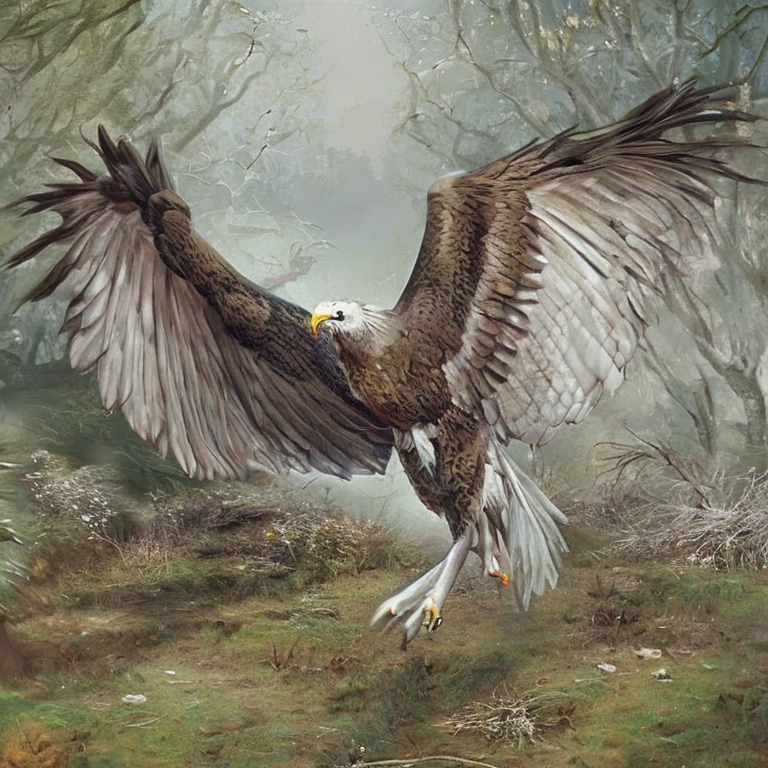}
      } \qquad
 \caption{Example images for the prompt ``a photo of a Hippogriffus hippogriffus, a mythical creature with the body of a horse and the head and wings of an eagle''.}%
 \label{fig:hippogriffus}
\end{figure*}

\begin{figure*}[!ht] 
 \centering
 \subfloat[]{%
      \includegraphics[width=0.26\textwidth]{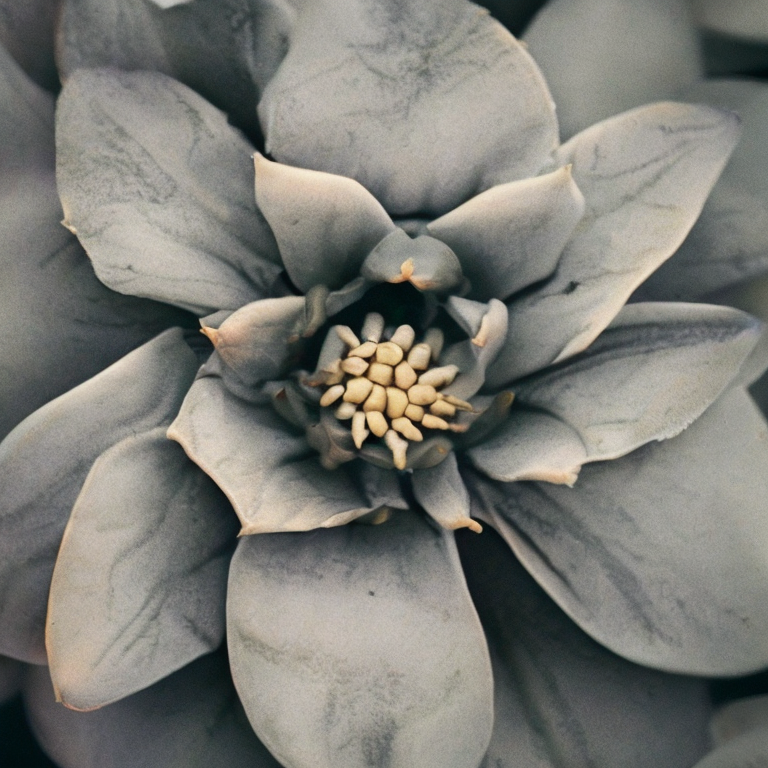}
      } \qquad
 \subfloat[]{
      \includegraphics[width=0.26\textwidth]{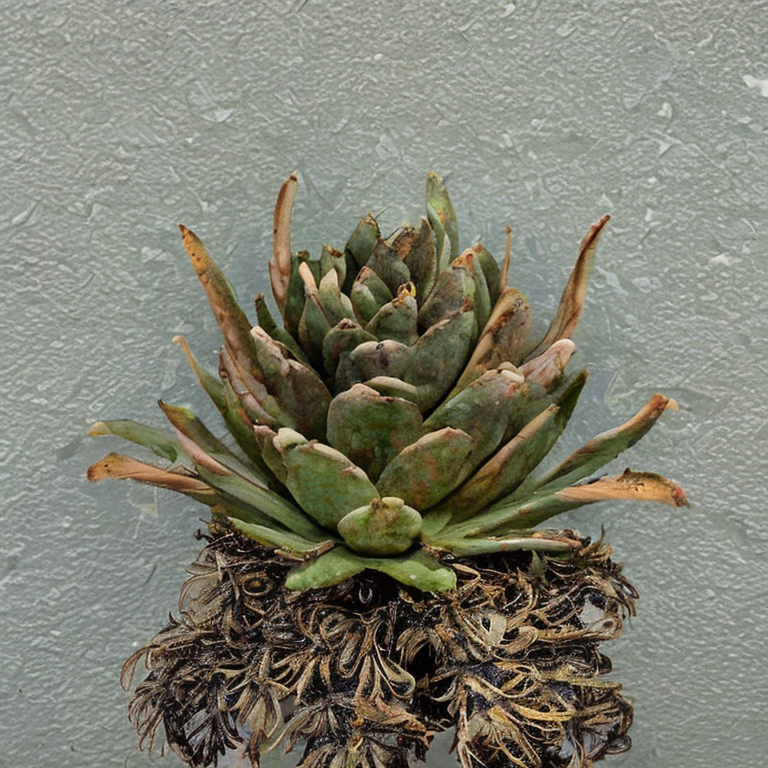} 
      } \qquad
 \subfloat[]{
      \includegraphics[width=0.26\textwidth]{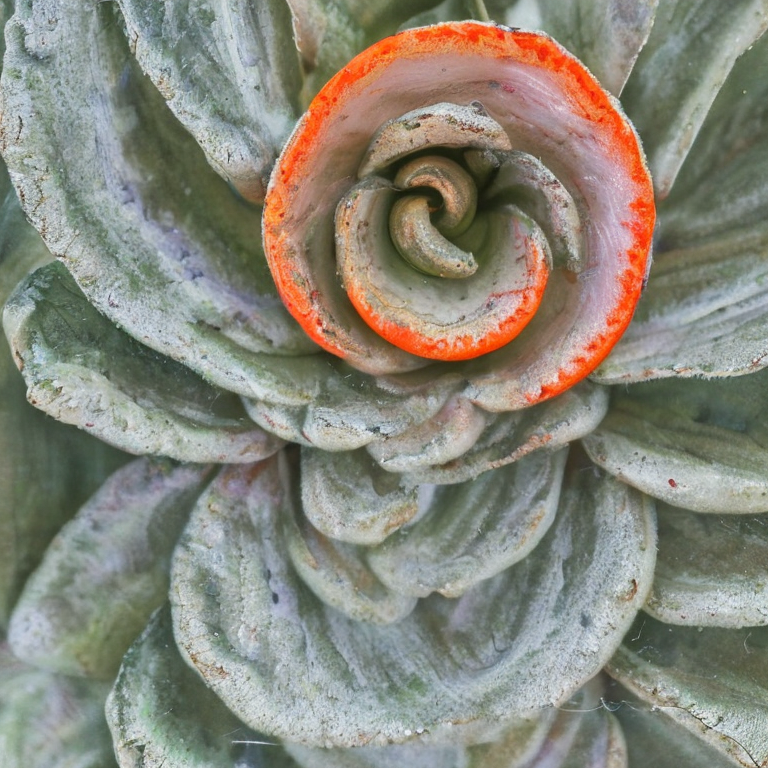} 
      } \qquad
 \caption{Example images for the prompt ``a photo of a Lion's ear, a succulent plant with fleshy leaves shaped like a lion's ear''.}%
 \label{fig:lions_ear}
\end{figure*}

\begin{figure*}[!ht] 
 \centering
 \subfloat[]{%
      \includegraphics[width=0.26\textwidth]{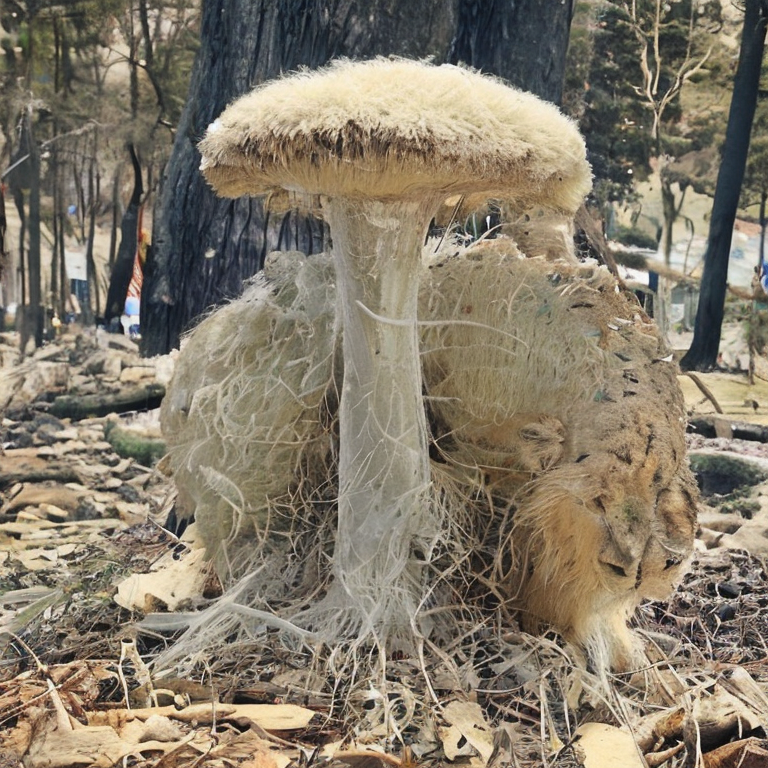}
      } \qquad
 \subfloat[]{
      \includegraphics[width=0.26\textwidth]{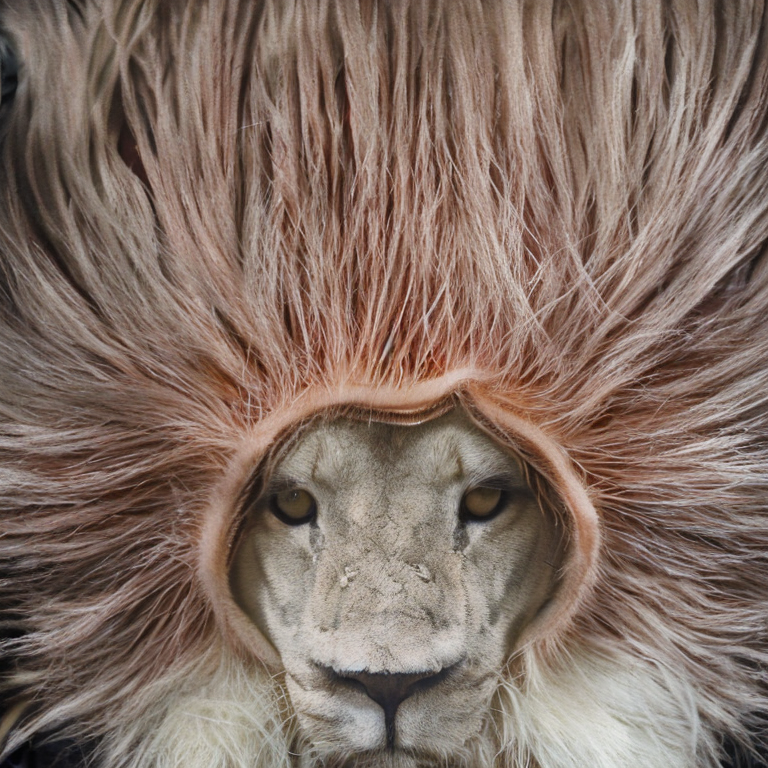} 
      } \qquad
 \subfloat[]{
      \includegraphics[width=0.26\textwidth]{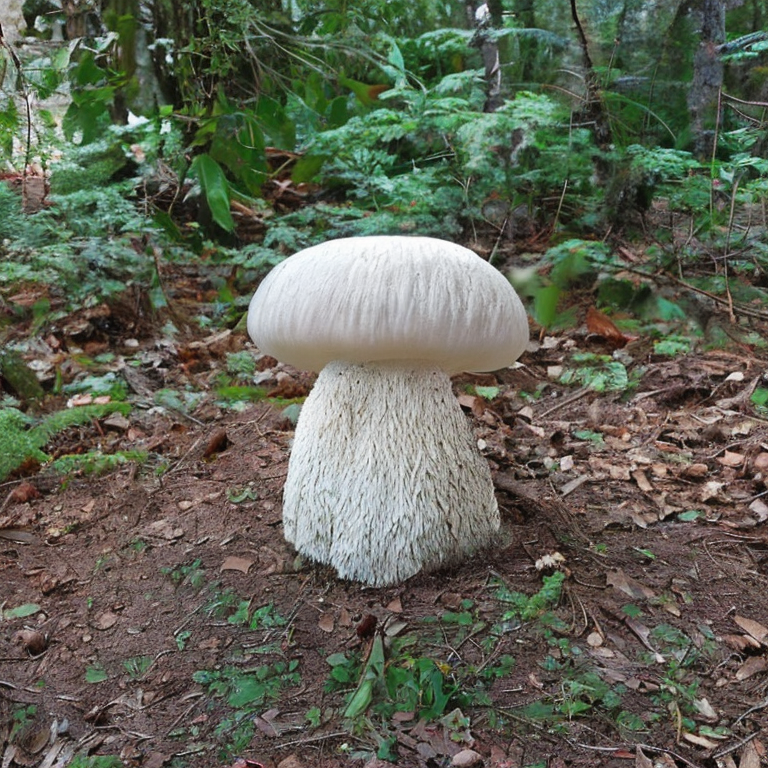}
      } \qquad
 \caption{Example images for the prompt ``a photo of a Lion's mane mushroom, a large, white mushroom with a distinctive lion's mane-like appearance''.}%
 \label{fig:lions_mane_mushroom}
\end{figure*}

We provide a few examples of images generated with stable-diffusion-2-1-base. 
Even with a low guidance scale, results are visually appealing, e.g. Figure \ref{fig:cod} and Figure \ref{fig:lion}. 
We observe that images are less realistic and homogeneous in the case of unusual or mythical objects, such as a hippogriffus in Figure \ref{fig:hippogriffus}. In this example, the generated image contains an animal that looks like an eagle, although the prompt specified that it should resemble that of a horse. 
Making prompts very descriptive (thus, very long) tends to be inefficient, as we observe that the beginning of the prompt has the greatest effect on the output (at least for our choice of tokenizer and textual encoder). 
We remark that natural species denoted by an ambiguous name such as ``lion's mane mushroom'' (a type of mushroom, Figure \ref{fig:lions_mane_mushroom}) or ``lion's ear'' (a type of succulent plant, Figure \ref{fig:lions_ear}) may produce semantically irrelevant pictures. 
In Figure \ref{fig:lions_ear} we see that even if the depicted object is a plant, there are glimpses of orange color in each picture, indicating a kind of ``semantic leakage'' between the lion concept (orange color), and the plant concept (plant shape). This phenomenon is particularly visible in Figure \ref{fig:lions_mane_mushroom}, where the mushrooms have something of a furry lion's mane - or even an entire lion's head, although the prompt specifies that the object is a mushroom.

\subsection{Comparison of images}
\begin{figure*}[ht] 
 \centering
 \subfloat[iNat1k]{
      \includegraphics[width=0.20\textwidth]{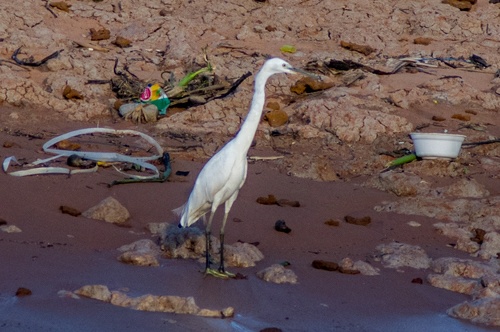} 
      }
 \subfloat[iNat1k]{
      \includegraphics[width=0.20\textwidth]{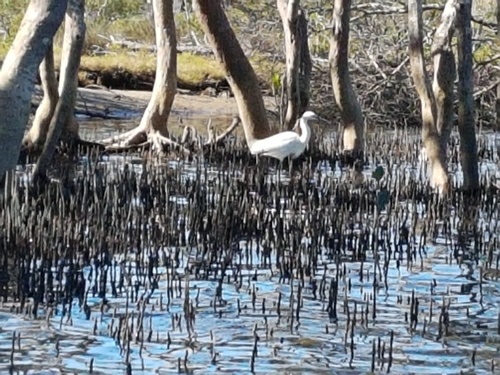}   
      }
 \subfloat[iNat1k]{
      \includegraphics[width=0.20\textwidth]{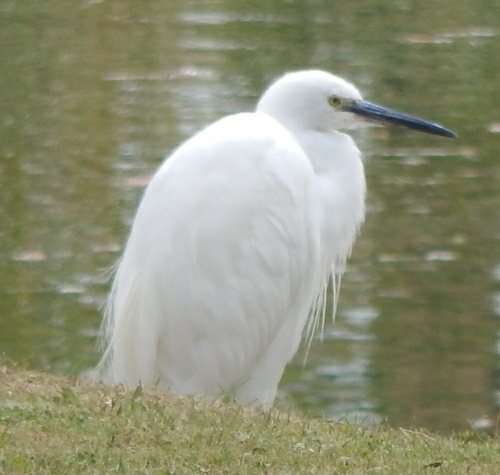} 
      }
 \subfloat[iNat1k]{
      \includegraphics[width=0.20\textwidth]{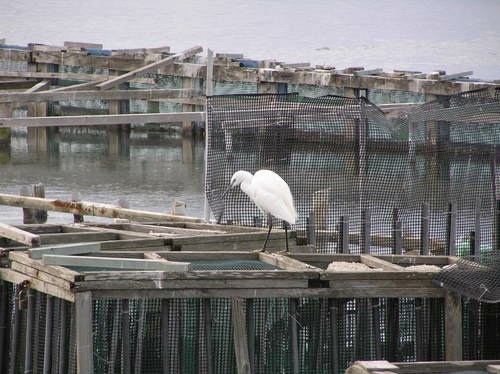} 
      } \\
 \subfloat[SimuGen]{%
      \includegraphics[width=0.20\textwidth]{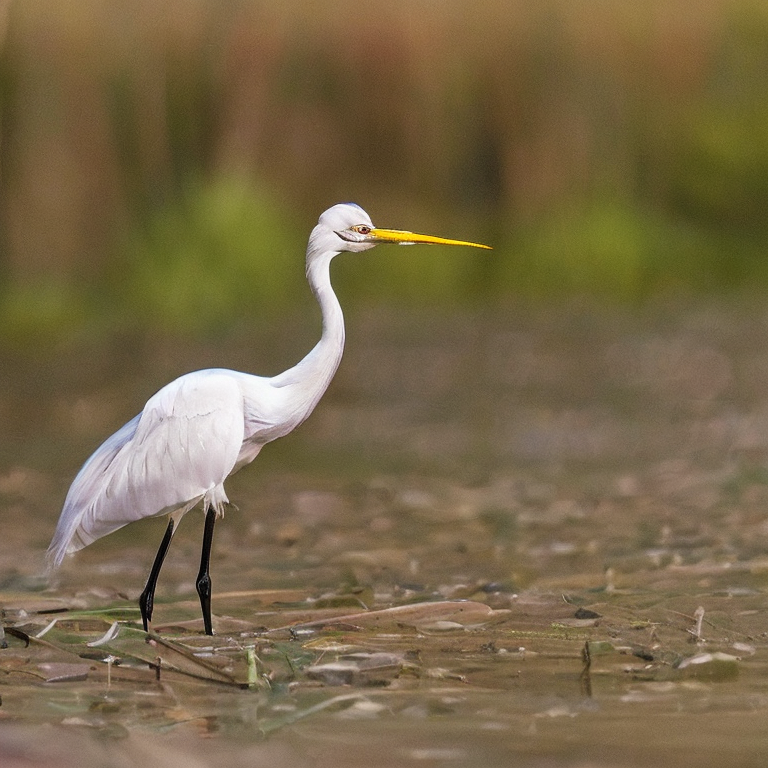} 
      } 
 \subfloat[SimuGen]{
      \includegraphics[width=0.20\textwidth]{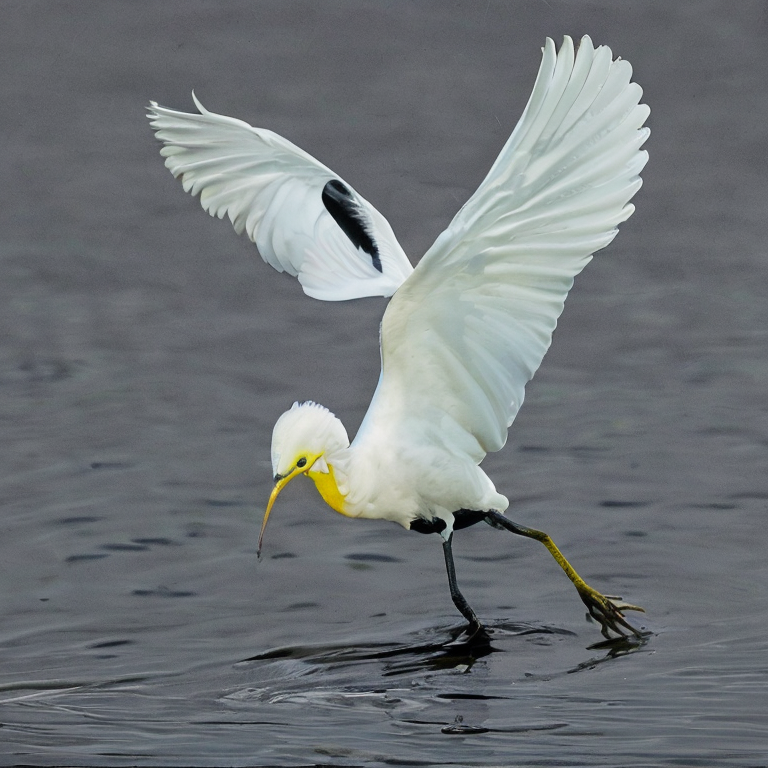} 
      } 
 \subfloat[SimuGen]{
      \includegraphics[width=0.20\textwidth]{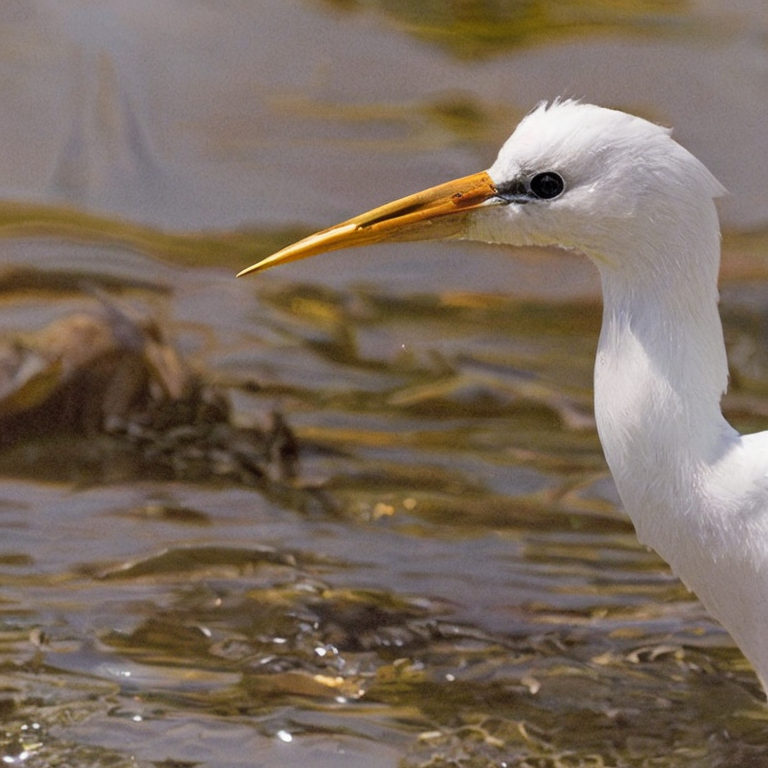} 
      } \qquad
 \subfloat[SimuGen]{
      \includegraphics[width=0.20\textwidth]{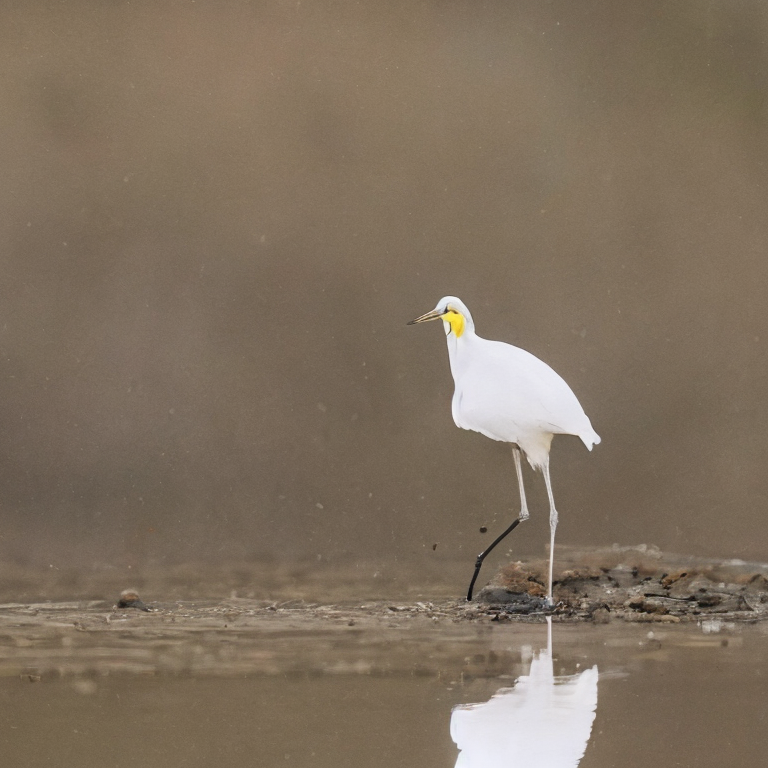}
      } \\
 \subfloat[Proxy21k]{%
      \includegraphics[width=0.20\textwidth]{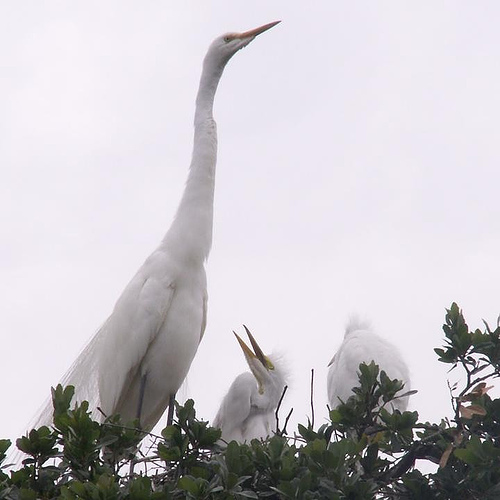} 
      }
 \subfloat[Proxy21k]{
      \includegraphics[width=0.20\textwidth]{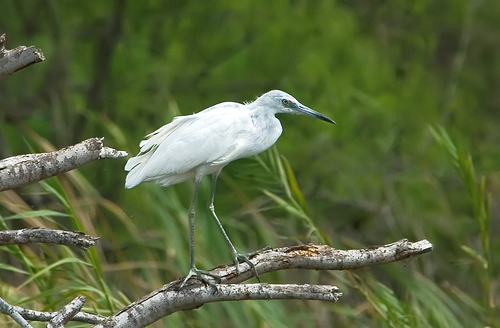} 
      }
 \subfloat[Proxy21k]{
      \includegraphics[width=0.20\textwidth]{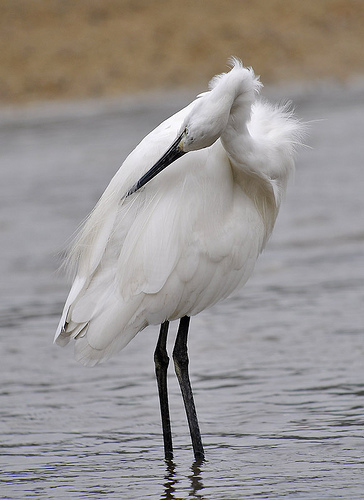} 
      }
 \subfloat[Proxy21k]{
      \includegraphics[width=0.20\textwidth]{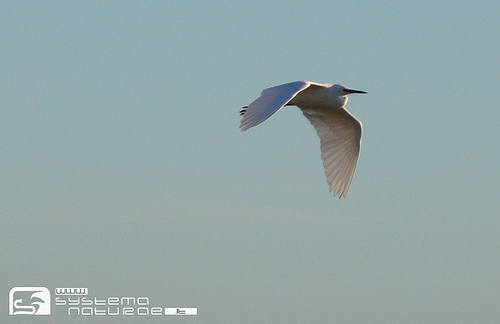} 
      }
\caption{Comparison of images from a class covered by the three versions of iNat1k dataset. (a - d) Examples of images from the original iNat1k, for class $04374\_$Animalia$\_$Chordata$\_$Aves$\_$Pelecaniformes$\_$Ardeidae$\_$Egretta$\_$garzetta. (e - h) Examples for the images obtained using stable-diffusion-2-1-base with the prompt ``a photo of an egretta garzetta, a small white egret with a yellow bill and black legs" and different random seeds. (i-l) Examples for the ImageNet class n02009508 named ``little egret, Egretta garzetta".} \label{fig:egretta}
\end{figure*}

In Figure \ref{fig:egretta} we show example images of the class ``egretta garzetta'' (a kind of bird) for iNat1k, and the simulated datasets obtained with SimuGen and ProxyIN1k. 
We can see that images from iNat1k and ImageNet21k have more diverse backgrounds and poses. This is due to the fact that the pictures were taken in the wild. Images generated by stable-diffusion-2-1-base also show of a diversity of photo framing and backgrounds, but these backgrounds are less detailed and tend to be blurry. Thus, the object is easier to spot in the image. A closer look at the synthetic pictures reveals slight anatomic anomalies: an abnormally pointed beak, a strange folding of the wings and an upside-down leg. 
Finally, we note that pictures taken by humans come at various resolutions, whereas synthetic images all come in the same size.

\subsection{Preliminary experiments with geometric transformations to simulate new classes}
\begin{figure*}[!htb]
\centering
\subfloat[Linear combination]{%
      \includegraphics[width=0.5\textwidth]{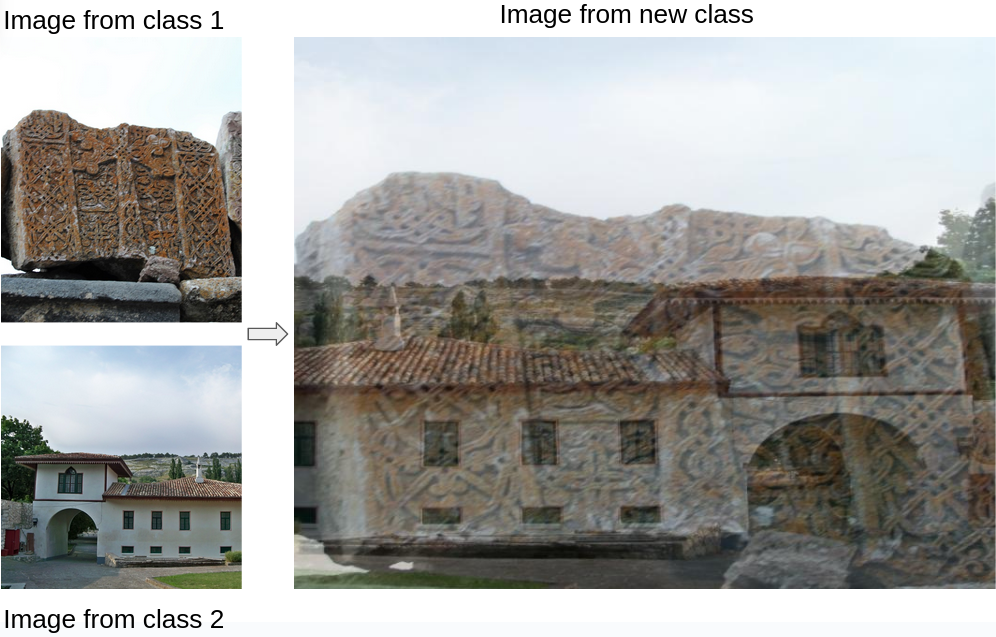} 
      \label{fig:half_mix}
      } \qquad
\subfloat[Minimum]{
    \includegraphics[width=0.5\linewidth]{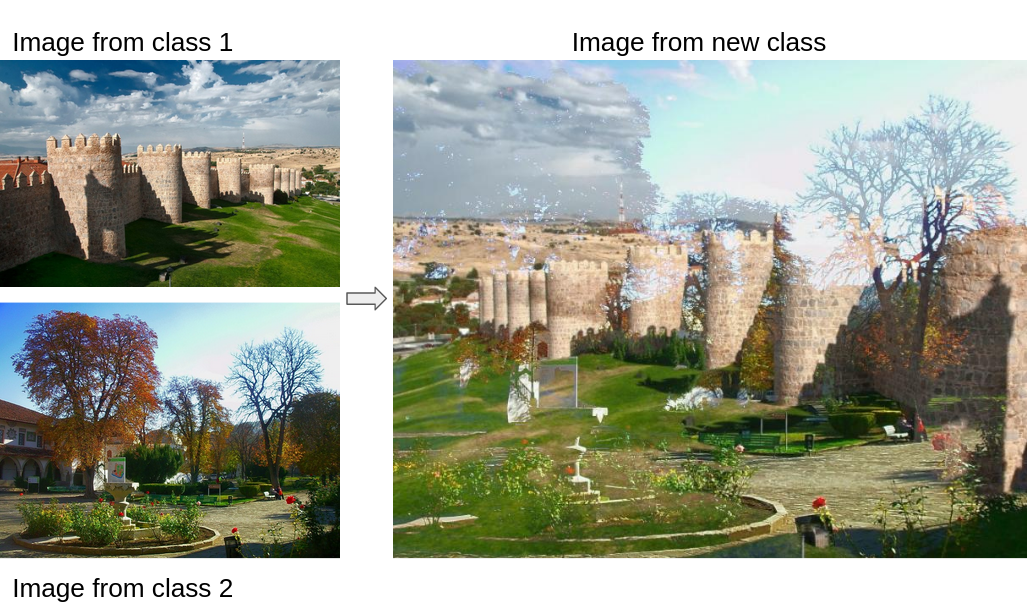}
    \label{fig:min_mix}
    } \qquad
\subfloat[HMix]{
    \includegraphics[width=0.5\linewidth]{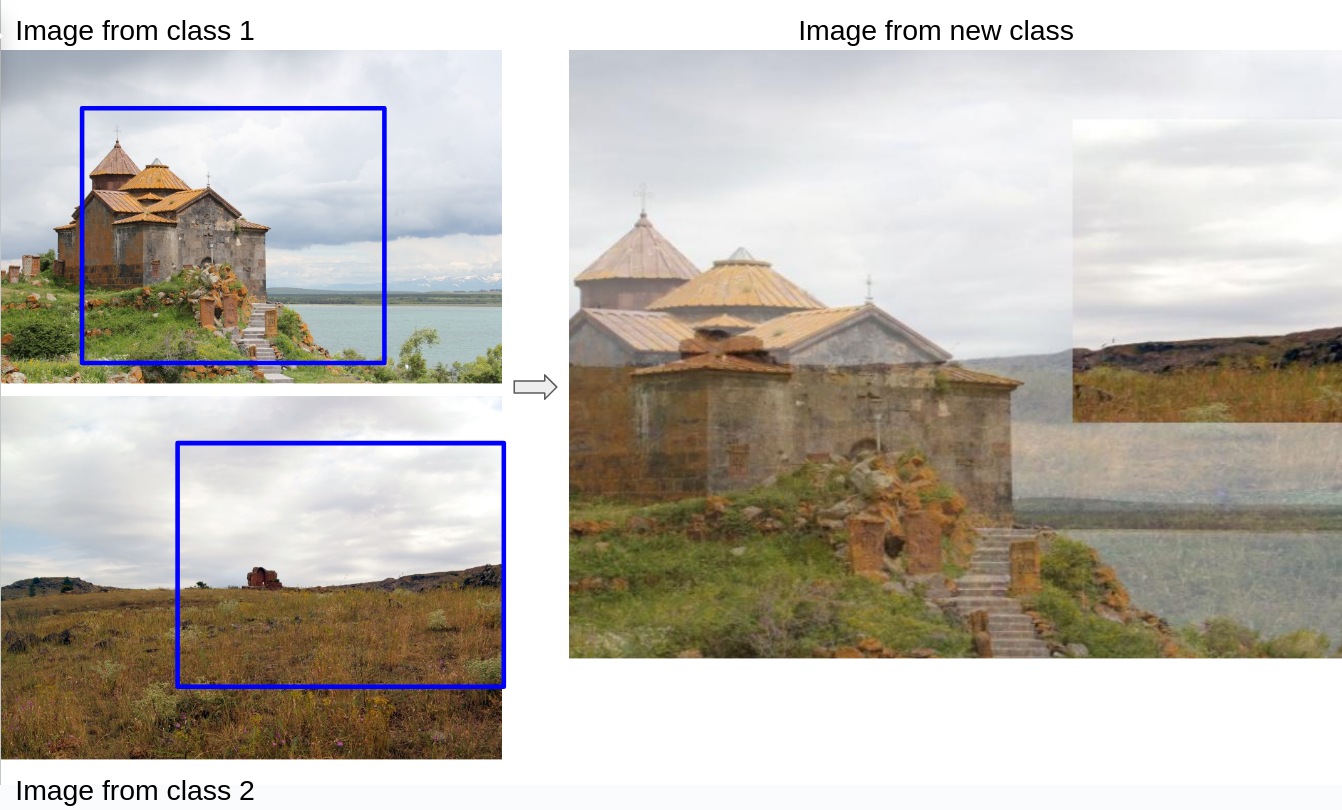}
    \label{fig:hmix}
    } \qquad
\subfloat[GMix]{
    \includegraphics[width=0.5\linewidth]{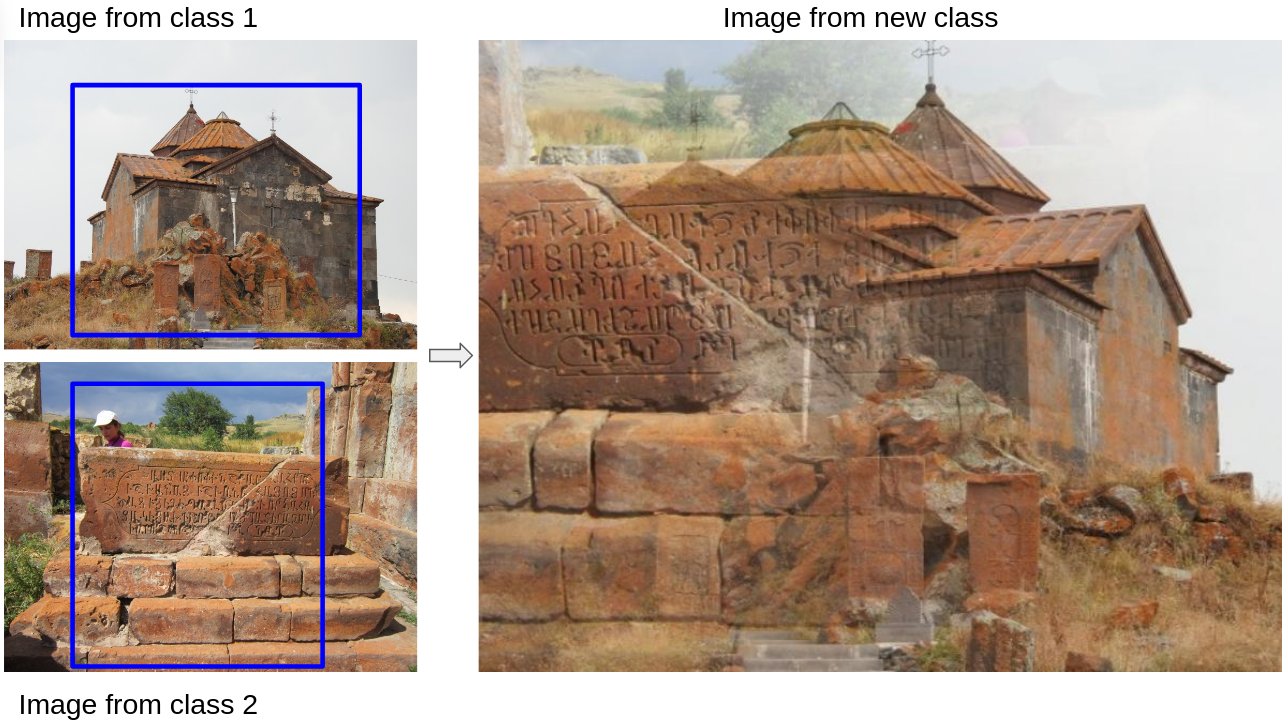}
    \label{fig:gmix}
    }
\caption{Examples of geometric augmentations for simulating new classes. Images from Land1k.} \label{fig:geometric_aug}
\end{figure*}

Before simulating additional classes by prompting a generative model, we experimented with simple pixel-wise transformations. We used Land1k in these preliminary experiments.
The principle of these geometric transformations is illustrated in Figure \ref{fig:geometric_aug}:
\begin{itemize}
    \item Figure \ref{fig:half_mix}: Convex pixel to pixel combination of two images
$I_{new} = \lambda I_1 + (1 - \lambda) I_2$, where $\lambda \in [0.4, 0.6]$.
    \item Figure \ref{fig:min_mix} Pixel-wise minimum (or maximum).
    \item Figure \ref{fig:hmix} Hybrid MixUp and CutMix (HMix) as defined in \cite{park2022unified}. 
    \item Figure \ref{fig:gmix} Gaussian MixUp as defined in \cite{park2022unified}.
    \item We also used rotations of 90, 180 and 270 degrees, but this is an issue when using algorithms such as SDC or PASS because these algorithms already use rotations to augment the training set.
\end{itemize}
We found out that the simulated stream of data was too far from the real data distribution to obtain relevant recommendations. 
DFCIL experiments on simulated data gave very low accuracy with very low variation from one algorithm to another, so the rankings of DFCIL algorithms were meaningless.  
We note that classic mix-up (convex pixelwise combination of two images) gave a better basis for recommendations than rotation but still it performed worse than any of the fixed baselines showed in the main results of the article, and of course than Proxy21k and SimuGen. 

%% file: sec_supp/3_efcil.tex
\newpage
\section{DFCIL experiments}

\subsection{DFCIL algorithms}
Our main experiments feature six competitive algorithms, namely FeCAM, BSIL, FeTrIL, DSLDA, PlaStIL and NCM.
Additionally, we performed preliminary experiments on Land1k with other fine-tuning based methods, namely SDC\cite{sdc_2020}, IL2A\cite{zhu2021class}, PASS~\cite{zhu2021pass} and SSRE~\cite{zhu2022self} but these algorithms were not recommended in any of the settings we tested. 
More precisely, SSRE had the best average performance of the four, but was always outperformed either by BSIL~\cite{jodelet2021balanced}, the fine-tuning-based algorithms included in our main evaluation, or by FeTrIL \cite{petit_2023_WACV}. These findings are aligned with those from the authors of FeTrIL\cite{petit_2023_WACV}, who show that FeTrIL outperforms SSRE in scenarios with 50$\%$ of the classes in the initial step. It is also coherent with the evaluation of PlaStIL\cite{petit2023plastil}, where PlaStIL outperforms PASS in scenarios with an equal number of classes in each step. 
Note that FeTrIL itself is outperformed by FeCAM (with a unique covariance matrix) in most settings. 
We also tried ABD \cite{smith2021always} but its model inversion component that creates synthetic images from past classes did not scale to 1000-class datasets. 

The computational cost of BSIL is comparable to or lower than that of the other fine-tuning-based methods we tested. For example, contrarily to SDC (or PASS) it does not require adding rotated images in each batch of the initial model (or of each model for PASS) to emulate additional classes during training. Contrarily to IL2A it does not introduce a data augmentation component. 
Finally, we remark that fine-tuning based algorithms tend to be more computationally intensive than fixed-representation-based algorithms. They also require GPU for the entire training, whereas FeTrIL, DSLDA and NCM can run on CPU once features have been computed. Regarding FeCAM, we ran it on GPU to accelerate the matrix inversion step. 
We also note that the same feature extractor can be trained once for all fixed-representation based methods to accelerate the DFCIL experiments.

The selection of the six algorithms used in our main experiments is driven by their complementary performance in the different incremental learning scenarios we tested, as well as their reasonable computational cost, which is also of practical interest.

We consider the models trained with the six selected algorithms after $t$ steps, for a total of $N_t$ classes.
At inference, all methods require a feature extractor, here a ResNet18~\cite{he2016_resnet} network containing around 10 million parameters. Features have a size $d$, here 512.
NCM needs the class prototypes of each class, i.e. it stores $d \cdot N_t$ values (at most $512 \cdot 10^3$ values in our experiments).
DSLDA and FeCAM each store class prototypes and a covariance matrix of size $d^2$, accounting for $512\cdot 10^3 + 262 \cdot 10^3$ values at most. Note that FeCAM version with one covariance matrix per class would result in $N_t \cdot d \cdot (d + 1)$ values instead of $d \cdot (N_t + 1)$ values. 
FeTrIL stores the parameters of $N_t$ linear SVCs i.e. $(d+1) \cdot N_t$ values ($513 \cdot 10^3$ values at most) and uses class prototypes to simulate the distribution of past classes ($512 \cdot 10^3$ values at most).
The linear classifier of BSIL contains $(d+1) \cdot N_t$ parameters.
PlaStIL is the most costly algorithm of the six in terms of storage, as it keeps multiple network heads in the limit of twice the number of parameters of the original neural network, as well as linear SVCs for each head, so around $23 \cdot 10^6$ values in total. 
PlaStIL was designed to have the same memory requirements \textit{at training time} as algorithms using knowledge distillation i.e. twice the model size, but as it needs to compute the embedding of an image for each model head at inference, this translates into an additional memory requirement at inference. Note that in the version used here, there are 4 heads. 
Nevertheless, it remains of the same order of magnitude as the other algorithms and an order of magnitude less expensive than FeCAM with one covariance matrix per class. 

\subsection{Implementation of DFCIL algorithms}
DFCIL algorithms are implemented using PyTorch implementation of the ResNet18 architecture\footnote{\url{https://github.com/pytorch/vision/blob/main/torchvision/models/resnet.py}}. Algorithm implementation is adapted from the following repositories.
\begin{itemize}
    \item DSLDA \cite{hayes2020_deepslda}: \url{https://github.com/tyler-hayes/Deep_SLDA}
    \item FeTrIL \cite{petit_2023_WACV}: \url{https://github.com/GregoirePetit/FeTrIL}
    \item BSIL \cite{jodelet2021balanced}: \url{https://github.com/hshustc/CVPR19_Incremental_Learning}.
    \item FeCAM \cite{goswami2024fecam}: \url{https://github.com/dipamgoswami/FeCAM}
    \item PlaStIL \cite{petit2023plastil}: \url{https://github.com/GregoirePetit/PlaStIL}
\end{itemize}
The implementation of NCM with cosine distance is derived from the code of FeCAM.

For a given dataset and set of initial classes, all algorithms share a common initial model trained on the initial classes. 
For the algorithms which use a fixed feature extractor (NCM, DSLDA, FeTrIL and FeCAM), the weights of the initial model trained following BSIL implementation are kept frozen and used to extract features of test samples.
Extracted features are then common to these four methods.

In the following, we provide key hyperparameters for each algorithm.

\textbf{DSLDA} The value of the shrinkage parameter before matrix inversion is $0.0001$.

\textbf{FeTrIL} Linear SVCs are trained using \texttt{scikit-learn} \cite{fabian2012_scikitlearn} implementation with $C = 1.0$ and $toler = 0.0001$.

\textbf{BSIL} The number of prototypes is set to 0. The number of simulated past samples in the balanced softmax loss is set to three percent of the number of training samples of the class.

Depending on the incremental learning scenario, we train each model for the following number of epochs with a batch size of $128$ and an SGD optimizer:  
\begin{itemize}
    \item[$-$] $20 + 49 \cdot 20$ classes : $60$ epochs for the initial model then $50$ epochs per incremental step.
    \item[$-$] $100 + 9\cdot50$, $200 + 4\cdot200$, $100 + 5\cdot100$ classes : $90$ then $60$ epochs.
    \item[$-$] $500 + 5\cdot100$ classes, $500 + 10\cdot 50$ classes : $120$ then $90$ epochs.
\end{itemize}
The initial learning rate is set to $0.1$, and the momentum is set to $0.9$.
The learning rate is reduced after $1/3$ and $2/3$ of the total number of epochs by a factor of $0.1$. The weight decay coefficient is $0.0001$.

\textbf{FeCAM}
For a fair comparison with the other algorithms in terms of memory requirements at inference time, we use the version of FeCAM with one common covariance matrix for all seen classes. Best results were obtained with an initial covariance shrinkage of 10.0 ($\gamma_1 = 10.0$, $\gamma_2 = 10.0$) and no covariance shrinkage in the incremental steps ($\gamma_1 = 0.0$, $\gamma_2 = 0.0$).
In preliminary experiments, we observed that when the initial model learns from 500 classes, FeCAM with a common covariance matrix performs almost the same as FeCAM with one covariance matrix per class.

\textbf{PlaStIL}
We use the version of PlaStIL which updates the last convolutional block of ResNet18 and combines four fine-tuned heads with linear SVCs. We use the same hyperparameters for SVCs as in FeTrIL. Fine-tuning of the heads is performed similarly as in BSIL, but with an initial learning rate of $0.01$ instead of $0.1$.

\textbf{NCM}
The prototype of a given class is obtained by averaging the feature vectors of the training samples belonging to this class. 
The nearest class mean classifier is based on the cosine distance to the class prototypes.

%% file: sec_supp/4_results.tex
\subsection{Detailed results for the main experiments}

\begin{figure*}[ht] 
\centering
 \subfloat[Best-performing algorithm] {
    \includegraphics[width=0.4\linewidth]{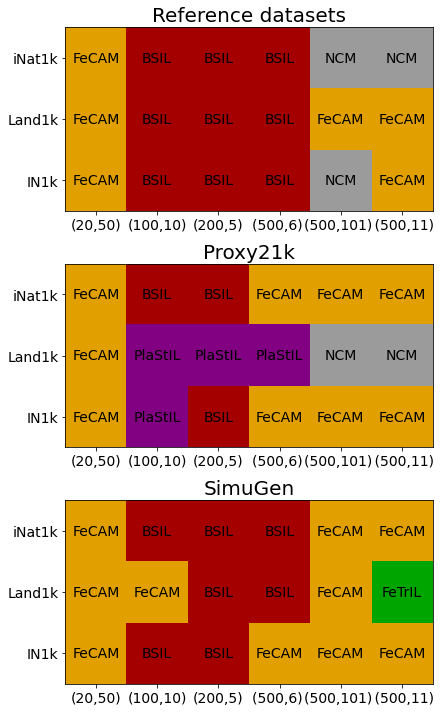} 
    \label{fig:heatmap_best}
    } \qquad
 \subfloat[Lowest-performing algorithm] {
    \includegraphics[width=0.4\linewidth]{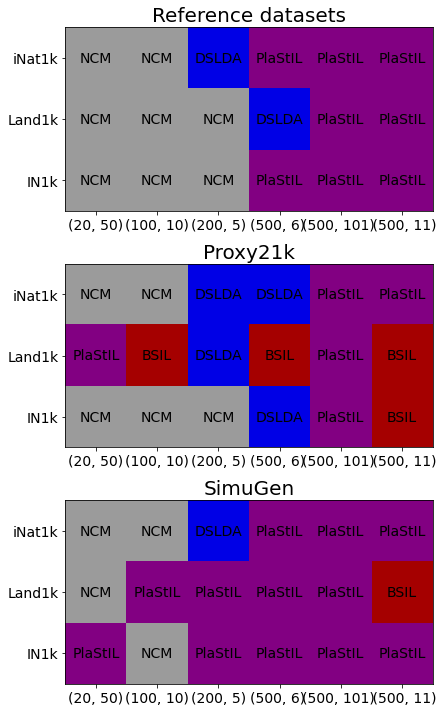} 
    \label{fig:heatmap_worst}
}
\caption{Highest- and lowest-algorithm for each dataset and each DFCIL scenario. The ranking is based on the average incremental accuracy of the DFCIL algorithms in the corresponding settings. DFCIL scenarios are described by the couple $(Card(P_1), T)$, i.e. the number of initial classes and the number of steps of the incremental process.}
\label{fig:heatmap_ranking}
\end{figure*}

The interest of recommending an algorithm depending on the dataset and on the scenario is illustrated by Figure \ref{fig:heatmap_ranking}. 
It shows the highest- and lowest-performing algorithm in terms of average incremental accuracy for each dataset in each incremental setting. 
The same algorithm, e.g. BSIL, may perform best in one setting and worst in another.
We note that in some cases, when the initial feature extractor has been trained on 500 classes, NCM has the highest average accuracy despite its simplicity. This echoes the findings of \cite{janson2022simple} in the case of pre-trained models. Nevertheless, it is the best performing only by a small margin (see Table \ref{tab_results_reco_supp}).

\begin{figure*}[ht] %
\centering
 \subfloat[iNat1k with $\text{Card}(P_1)=20$ and $T=50$]{%
      \includegraphics[width=0.82\textwidth]{img/inat1k_b20_t49.png} \label{fig:inat1k_b20_t49}
      } \qquad
 \subfloat[Land1k with $\text{card}(P_1) = 100$ and $T = 10$]{%
      \includegraphics[width=0.82\textwidth]{img/land1k_b100_t9.png}   \label{fig:land1k_b100_t9}
      } \qquad
 \subfloat[Land1k with $\text{card}(P_1) = 200$ and $T = 5$]{%
      \includegraphics[width=0.82\textwidth]{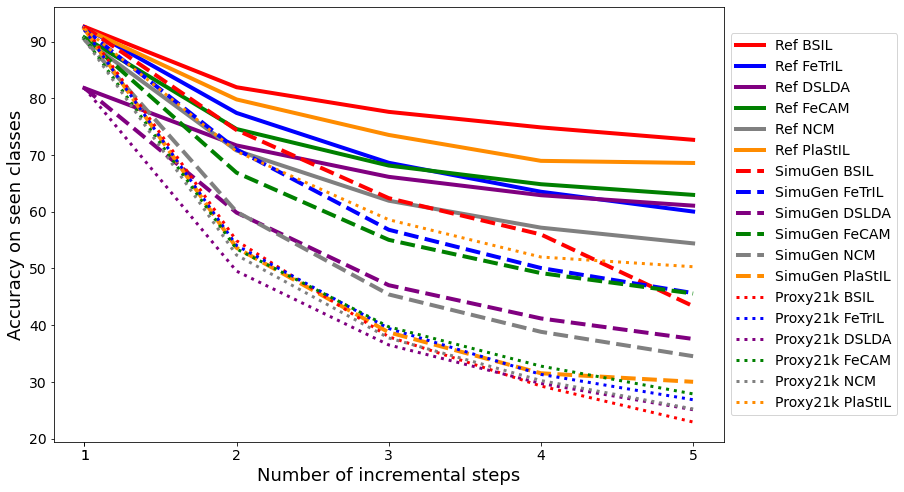}   \label{fig:land1k_b200_t4}
      }
\caption{Detailed incremental accuracy for three scenarios with an equal number of classes per step.}%
\label{fig:learning_curves_equi}
\end{figure*}

\begin{figure*}[ht] 
\centering
\subfloat[IN1k with $\text{Card}(P_1)=500$ and $T=6$)]{%
      \includegraphics[width=0.82\textwidth]{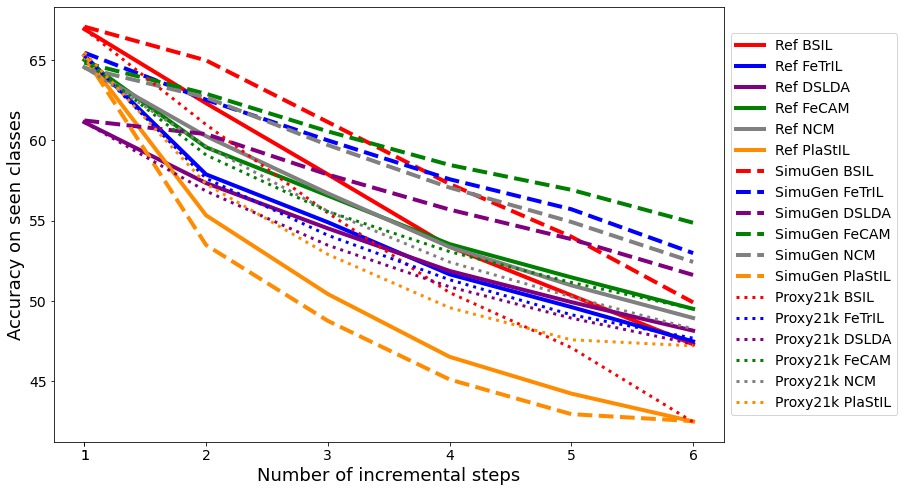} 
      \label{fig:inat1k_b500_t5}
      } \qquad
\subfloat[iNat1k with $\text{Card}(P_1)=500$ and $T=11$]{%
      \includegraphics[width=0.82\textwidth]{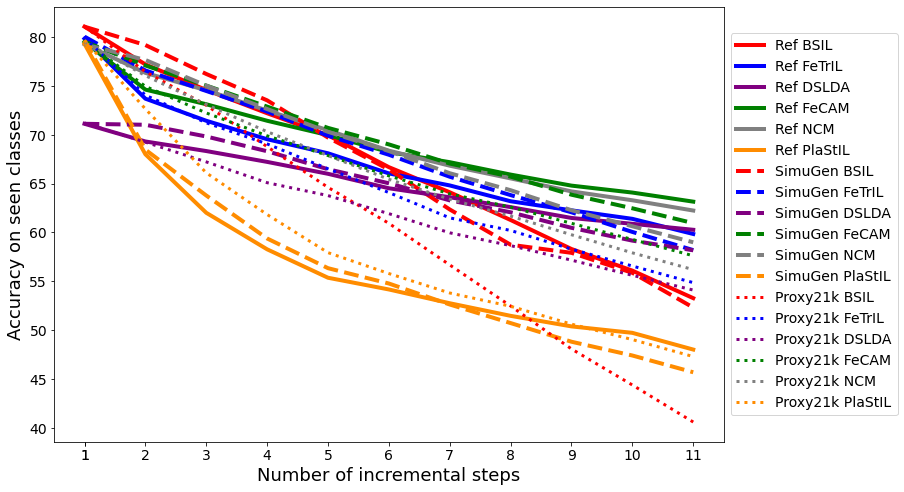} 
      \label{fig:inat1k_b500_t10}
      } \qquad
\subfloat[IN1k with $\text{Card}(P_1)=500$ and $T=101$]{%
      \includegraphics[width=0.82\textwidth]{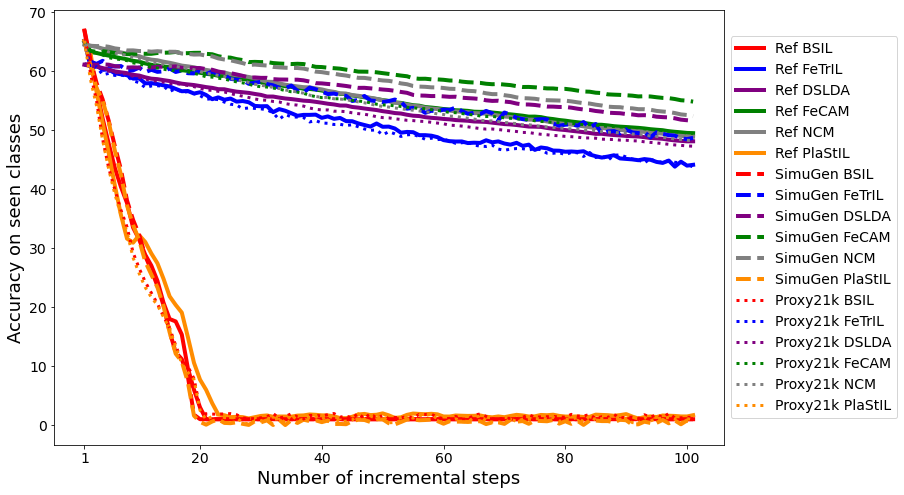} 
      \label{fig:ilsvrc_b500_t100}
      }
\caption{Detailed incremental accuracy for three scenarios with half of the classes in the initial step, and the other half of the classes equally distributed among the incremental steps.}%
\label{fig:learning_curves_semi}
\end{figure*}

We provide examples of learning curves for various datasets and scenarios in Figures \ref{fig:learning_curves_equi} and \ref{fig:learning_curves_semi}. 
In the most challenging scenario where the thousand classes are split in fifty steps of twenty classes each (Figure \ref{fig:inat1k_b20_t49}), algorithms with a fixed feature extractor and a dedicated classifier learning module (FeCAM, FeTrIL and DSLDA) perform best, despite the fact that the initial feature extractor was not trained on a large dataset. NCM exhibits the lowest performance, indicating that the features extractor is not able to produce features which can be easily separated in the embedding space. BSIL and PlaStIL are challenged by the multiple iterations of the fine-tuning procedure, even if knowledge distillation is used by BSIL or if most of the parameters are kept fixed for PlaStIL.

In scenarios where each step contains an equal number of classes and the initial model is trained on enough classes to transfer well to the next classes, BSIL significantly outperforms the other algorithms, followed by PlaStIL and FeCAM. The accuracy of DSLDA is quite stable, but the LDA classifier seems less well adapted for separating classes in this embedding space than the classifiers of the other algorithms. 

Regarding scenarios where the initial model is trained on half of the classes (Figure \ref{fig:learning_curves_semi}), algorithms based on a fixed representation benefit from the increased transferability of the feature extractor and exhibit very stable performances. BSIL performs well too, unless incremental steps are numerous and require fine-tuning the model on a small number of classes at each time, causing a rapid forgetting of the representation learned initially.
PlaStIL is the lowest-performing algorithm in these scenarios, showing the limit of its fine-tuning mechanism with multiple heads.    

Individual results in terms of average incremental accuracy are provided in Table \ref{tab:detailed_logs}. For each reference dataset and each incremental setting $(Card(P_1), T)$, we provide the average incremental accuracy of the best performing DFCIL algorithm, and the difference between the performance of the other algorithms and the performance of this algorithm.  

\begingroup
\setlength{\tabcolsep}{4pt} 
\renewcommand{\arraystretch}{1.1} 
\begin{table}[t]
\caption{Detailed results: $\rho_{ref}$ is the best average incremental accuracy for each reference dataset in each incremental learning scenario (oracle). $\Delta_{gen}^T$ (or $\Delta_{proxy}^T$) denotes the performance gap between the algorithm recommended after $T$ simulation steps on the data stream simulated following SimuGen approach (or following Proxy21k, respectively). $\Delta_m$ denotes the performance gap between the algorithm recommended by the oracle and algorithm $m$. The algorithms PlaStIL, BSIL, NCM, DSLDA, FeTrIL and FeCAM are abbreviated by $P$, $B$, $N$, $D$, $F$ and $Fc$, respectively.} \label{tab:detailed_logs}
\resizebox{0.990\linewidth}{!}
{
\begin{tabular}{|c|l|l|ll|llllll|}
\hline
\multicolumn{1}{|l|}{Dataset} & $(Card(P_1), T)$ & $\rho_{ref}$ &$\Delta_{gen}^T$ & $\Delta_{proxy}^T$ & $\Delta_P$ & $\Delta_B$ & $\Delta_N$ & $\Delta_D$ & $\Delta_F$ & $\Delta_{Fc}$ \\ 
\hline
\multirow{6}{*}{IN1k}         &  (20, 50)   &  28.07  &  0.0  &  0.0  &  -9.06  &  -11.86  &  -12.82  &  -1.61  &  -2.39  &  0.0 \\
                              &  (100, 10)  &  44.05  &  0.0  &  -3.82  &  -3.82  &  0.0  &  -7.85  &  -3.08  &  -1.28  &  -0.94 \\
                              &  (200, 5)   &  54.64  &  0.0  &  0.0  &  -6.29  &  0.0  &  -9.27  &  -8.15  &  -5.8  &  -5.71 \\
                              &  (500, 6)   &  56.35  &  -0.41  &  -0.41  &  -5.65  &  0.0  &  -0.56  &  -2.54  &  -1.89  &  -0.41 \\
                              &  (500, 11)  &  55.72  &  0.0  &  0.0  &  -11.5  &  -7.24  &  -0.05  &  -2.01  &  -2.18  &  0.0 \\
                              &  (500, 101) &  55.58  &  -0.04  &  -0.04  &  -48.89  &  -48.41  &  0.0  &  -1.95  &  -4.45  &  -0.04  \\ 
                              \hline
\multirow{6}{*}{iNat1k}       &  (20, t50)  &  30.91  &  0.0  &  0.0  &  -10.34  &  -7.64  &  -13.4  &  -2.0  &  -2.52  &  0.0 \\
                              &  (100, 10)  &  57.59  &  0.0  &  0.0  &  -10.2  &  0.0  &  -15.04  &  -10.11  &  -7.11  &  -7.18  \\
                              &  (200, 5)   &  66.22  &  0.0  &  0.0  &  -7.6  &  0.0  &  -12.05  &  -12.25  &  -8.02  &  -8.5  \\
                              &  (500, 6)   &  71.55  &  0.0  &  -1.93  &  -6.93  &  0.0  &  -1.96  &  -6.41  &  -3.32  &  -1.93 \\
                              &  (500, 11)  &  69.41  &  -0.14  &  -0.14  &  -12.19  &  -2.62  &  0.0  &  -4.38  &  -2.13  &  -0.14 \\
                              &  (500, 101) &  69.26  &  -0.28  &  -0.28  &  -63.46  &  -62.98  &  0.0  &  -4.32  &  -3.76  &  -0.28 \\ 
                              \hline
\multirow{6}{*}{Land1k}       &  (20, 50)  &  46.39  &  0.0  &  0.0  &  -15.64  &  -15.13  &  -24.79  &  -3.31  &  -8.14  &  0.0  \\
                              &  (100, 10) &  70.02  &  -4.22  &  -4.97  &  -4.97  &  0.0  &  -15.74  &  -7.48  &  -7.38  &  -4.22  \\
                              &  (200, 5)  &  79.95  &  0.0  &  -3.3  &  -3.3  &  0.0  &  -12.99  &  -11.22  &  -7.53  &  -7.69  \\
                              & (500, 6)   &  85.28  &  0.0  &  -5.57  &  -5.57  &  0.0  &  -6.91  &  -9.59  &  -5.76  &  -6.18  \\
                              & (500, 11)  &  78.81  &  -0.63  &  -0.62  &  -5.51  &  -0.49  &  -0.62  &  -3.24  &  -0.63  &  0.0 \\
                              & (500, 101) &  78.55  &  0.0  &  -0.49  &  -68.41  &  -67.94  &  -0.49  &  -3.09  &  -2.68  &  0.0 \\ 
                              \hline
\end{tabular}
}
\end{table}
\endgroup

\begin{figure*}[ht] 
 \centering \vspace{-2em}
 \subfloat[(20,50)]{%
      \includegraphics[height=0.16\textheight]{img/timeline_b20_t49.png} \label{fig:timeline_b20_t49}  }  
      \hspace{0.3em}
 \subfloat[(100,10)]{%
      \includegraphics[height=0.16\textheight]{img/timeline_b100_t9.png} \label{fig:timeline_b100_t9} }      \hspace{0.3em}
 \subfloat[(200,5)]{%
      \includegraphics[height=0.16\textheight]{img/timeline_b200_t4.png} \label{fig:timeline_b200_t4} }     \hspace{0.3em}
 \subfloat[(500,6)]{%
      \includegraphics[height=0.16\textheight]{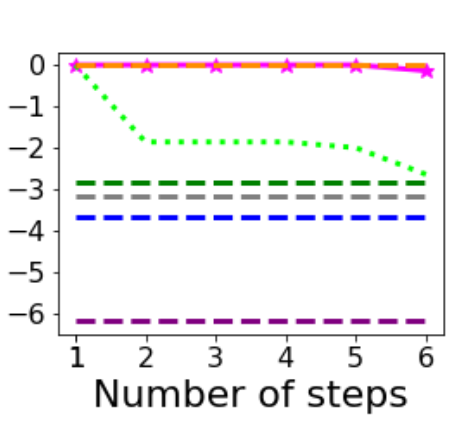} \label{fig:timeline_b500_t5} }     \hspace{0.3em}
 \subfloat[(500,11)]{%
      \includegraphics[height=0.16\textheight]{img/timeline_b500_10.png} \label{fig:timeline_b500_10} }     \hspace{0.3em}
 \subfloat[(500,101)]{%
      \includegraphics[height=0.16\textheight]{img/timeline_b500_t100.png} \label{fig:timeline_b500_t100}} 
 \caption{Detailed view of the performance gap between the oracle and the recommendation methods after simulating $t = 1, 2, \dots T$ incremental steps, for scenarios of the form ($Card(P_1)$, T). Results are averaged over the three datasets (IN1K, iNat1k, Land1k).}
  \label{fig_dynamics_supp}
  \vspace{-1.5em}
\end{figure*}

\textbf{Recommendation}
In Table \ref{tab_results_reco_supp} we provide additional results with a strategy consisting in running the DFCIL experiments on the simulated stream during $T/2$ steps out of $T$. 
We compare this strategy with the strategies already presented in the article, namely : 
\begin{itemize}
    \item running all DFCIL algorithms on the $T$ steps of the incremental process, 
    \item running all DFCIL algorithms on the 3 first steps only,
    \item running all DFCIL algorithms on 3 steps and then eliminating the lowest- performing candidate algorithm at each step until either $T$ is reached or only a single algorithm remains (``effi'').
\end{itemize} 
Running all experiments on all steps allows to get a maximum of information to make the recommendation. In Figure \ref{fig_dynamics_supp} we also provide in higher resolution the figures illustrating the dynamics of recommendations already provided in the article. 
The strategy with only $T/2$ simulation steps yields slightly less accurate recommendations than with $T$ simulation steps but is competitive. 
In the case of SimuGen, it even yields better results, because of an artifact in the scenario with 100 classes during 10 steps, visible in Figure \ref{fig:timeline_b100_t9}. The ``effi'' strategy strikes a balance between exploring all algorithms during the entire incremental process (T steps), or only at the beginning (3 steps).   

If we use the simulated stream of data to predict the second-best method, the algorithms provided by the oracle have an average incremental accuracy of $58.51\%$ on average, and algorithms recommended on the basis of SimuGen are $0.39\%$ below the oracle, whereas recommendations based on Proxy21k are $0.74\%$ below the oracle. 
Finally, if the user wants to use our recommendation method to decide which algorithm to discard in priority, SimuGen and Proxy21k are relevant too. 
In our experiments, the lowest performing algorithms have an average incremental accuracy of $41.23\%$ on average, and predicting the lowest-performing algorithm using SimuGen (or Proxy21k) leads to an average incremental accuracy $2.01\%$ (or $2.73\%$) away from the oracle on average, indicating that a low-performing algorithm is indeed picked in most of the settings, as showed in Figure \ref{fig:heatmap_worst}.

Finally, if we only have access to $k$ of the 6 algorithms used in our main experiments, Table \ref{tab:ablation_reco} shows that the recommendations remain relevant, in particular following the data simulation approach of SimuGen.

\begingroup
\setlength{\tabcolsep}{3pt} 
\renewcommand{\arraystretch}{1.1} 
\begin{table}
\begin{center}
\caption{Performance gap between the average incremental accuracy of the methods proposed by the oracle ($\rho_{ref}$) and that of the algorithms recommended with different methods (results averaged over (i) the three datasets (ii) the six DFCIL settings of the form $(\text{card}(P_1), T)$, (iii) all settings). The three strongest fixed baselines are shown for comparison (FeCAM : Fc, FeTrIL: F, DSLDA: D). Gaps closer to zero are better.
}  \vspace{-0.6em} 
\label{tab_results_reco_supp}
\resizebox{0.990\linewidth}{!}
{
\begin{tabular}{|l|l|l|llll|llll|lll|} 
\hline
\multicolumn{3}{|l|}{}  & \multicolumn{11}{|l|}{  \color{white}Some more space \color{black} Accuracy gap for recommendation methods} \\ 
\cline{3-14}
\multicolumn{2}{|l|}{}  & $\rho_{ref}$ & $\Delta_{gen}^{T}$ & $\Delta_{gen}^{T/2}$ & $\Delta_{gen}^{effi}$ & $\Delta_{gen}^{3}$ & $\Delta_{proxy}^{T}$ & $\Delta_{proxy}^{T/2}$  & $\Delta_{proxy}^{effi}$ & $\Delta_{proxy}^{3}$ & $\Delta_D$ & $\Delta_F$ & $\Delta_{Fc}$ \\ \hline 
\multicolumn{1}{|l|}{\raisebox{-2.5\normalbaselineskip}[0pt][0pt]{\rotatebox[origin=c]{90}{Scenario}}} & 
                                                       (20, 50) & 35.12  &  0.0  &  0.0  &  -3.95  &  -3.95  &  0.0  &  0.0  &  -2.55  &  -6.5  &  -2.31  &  -4.35  &  0.0  \\
\multicolumn{1}{|l|}{}                               & (100, 10) & 57.22  &  -1.41  &  0.0  &  0.0  &  0.0  &  -2.93  &  -1.66  &  -1.66  &  -1.66  &  -6.89  &  -5.26  &  -4.11\\ 
\multicolumn{1}{|l|}{}                               & (200, 5) & 66.94  &  0.0  &  0.0  &  0.0  &  0.0  &  -1.1  &  -1.1  &  -1.1  &  -1.1  &  -10.54  &  -7.12  &  -7.3 \\
\multicolumn{1}{|l|}{}                               & (500, 5) & 71.06  &  -0.14  &  0.0  &  0.0  &  0.0  &  -2.64  &  -1.86  &  -1.99  &  -1.86  &  -6.18  &  -3.66  &  -2.84 \\
\multicolumn{1}{|l|}{}                               & (500, 11) & 67.98  &  -0.26  &  -1.08  &  -0.26  &  -3.5  &  -0.25  &  -0.22  &  -0.22  &  -1.05  &  -3.21  &  -1.65  &  -0.05 \\
\multicolumn{1}{|l|}{}                               & (500, 101) & 67.8  &  -0.11  &  -0.11  &  -0.16  &  -0.89  &  -0.27  &  -0.16  &  -0.16  &  -0.16  &  -3.12  &  -3.63  &  -0.11 \\ \hline
\multicolumn{1}{|l|}{\raisebox{-1.1\normalbaselineskip}[0pt][0pt]{\rotatebox[origin=c]{90}{Dataset}}}      
                                                     & IN1k  & 49.07  &  -0.08  &  -0.01  &  -1.98  &  -3.18  &  -0.71  &  -0.01  &  -0.08  &  -1.98  &  -3.22  &  -3.0  &  -1.18 \\
\multicolumn{1}{|l|}{}                               & iNat1k & 60.82  &  -0.07  &  -0.48  &  -0.02  &  -0.44  &  -0.39  &  0.0  &  -1.27  &  -1.71  &  -6.58  &  -4.48  &  -3.01 \\ 
\multicolumn{1}{|l|}{}                               & Land1k  & 73.17  &  -0.81  &  -0.1  &  -0.19  &  -0.55  &  -2.49  &  -2.49  &  -2.49  &  -2.47  &  -6.32  &  -5.35  &  -3.02\\
\hline
\multicolumn{2}{|l|}{Average}  & 61.02  & -0.32  &  -0.2  &  -0.73  &  -1.39  &  -1.2  &  -0.83  &  -1.28  &  -2.06  &  -5.37  &  -4.28  &  -2.4 \\ \hline
\end{tabular}
}
\end{center}
\vspace{-1.5em} 
\end{table}
\endgroup

\begingroup
\setlength{\tabcolsep}{3pt} 
\renewcommand{\arraystretch}{1.1} 
\begin{table}[ht] 
\begin{center}
\caption{Performance gap between the average incremental accuracy ($\rho_{ref}^k$) of the algorithm provided by the oracle and that of the algorithm recommended by simulating the data stream following SimuGen approach ($\Delta_{gen}^k$) or Proxy21k approach ($\Delta_{proxy}^k$), when $k$ candidate algorithms are available out of the six. Results are averaged over the three reference datasets and the six incremental learning scenarios showed in the main experiments. Gaps closer to zero are better.}  
\label{tab:ablation_reco}
\resizebox{0.60\linewidth}{!}
{
\begin{tabular}{l|llllll|l}
\cline{2-7}
  & \multicolumn{6}{c|}{Number of available algorithms $k$}  &   \\ 
\cline{2-8} 
 & 1 & 2 & 3 & 4 & 5 & 6 & \multicolumn{1}{l|}{average} \\ 
\hline
\multicolumn{1}{|l|}{$\rho_{ref}^k$} &  52.92 & 57.70 & 59.40 & 60.09 & 60.60 & 61.02 & \multicolumn{1}{l|}{58.62} \\ 
\hline
\multicolumn{1}{|l|}{$\Delta_{proxy}^k$}   &   0.0 & -0.67 & -0.82 & -0.98 & -1.12 & -1.20 & \multicolumn{1}{l|}{-0.80}  \\ 
\hline
\multicolumn{1}{|l|}{$\Delta_{gen}^k$}     &   0.0 & -0.32 & -0.28 & -0.29 & -0.32 & -0.32 & \multicolumn{1}{l|}{-0.26}  \\ 
\hline
\end{tabular}
}
\end{center}
\end{table}
\endgroup